%% file: main.tex
\documentclass[acmtog, authorversion]{acmart}

\usepackage{booktabs} %

\setcitestyle{square} %
\input{macros}

\usepackage{amssymb}
\usepackage{bm}
\usepackage{subcaption}
\setcopyright{acmcopyright}
\acmJournal{TOG}
\acmYear{2021} \acmVolume{1} \acmNumber{1} \acmArticle{1} \acmMonth{1} \acmPrice{15.00}\acmDOI{10.1145/3451340}

\usepackage[ruled]{algorithm2e} %

\SetAlFnt{\small}
\SetAlCapFnt{\small}
\SetAlCapNameFnt{\small}
\SetAlCapHSkip{0pt}

\begin{document}

\title{Learning Multimodal Affinities for Textual Editing in Images} 

\author{Or Perel}
\affiliation{%
\institution{Amazon Web Services}}
\email{orperel@amazon.com}

\author{Oron Anschel}
\affiliation{%
\institution{Amazon Web Services}}
\email{oronans@amazon.com}

\author{Omri Ben-Eliezer}
\affiliation{%
\institution{Harvard University$^{\dagger}$}}
\email{omribene@cmsa.fas.harvard.edu}

\author{Shai Mazor}
\affiliation{%
\institution{Amazon Web Services}}
\email{smazor@amazon.com}

\author{Hadar Averbuch-Elor}
\affiliation{%
\institution{Cornell-Tech, Cornell University$^{\dagger}$}}
\email{hadarelor@cornell.edu}

\begin{abstract}
\input{abstract}
\end{abstract}

\input{acm_classification_terms}

\input{figures/teaser/teaser.tex}

\maketitle

\input{intro}
  \input{related}

\input{overview}

 \input{words}
 \input{features}
 \input{optimization}
 \input{clustering}
\input{semisupervised}

\input{results}

\input{applications}

\input{conclusions}

\bibliographystyle{ACM-Reference-Format}
\bibliography{parsedoc}

\end{document}

%% file: macros.tex
\long\def\ignorethis#1{}

\usepackage{amsmath}

\usepackage{ amssymb }
\usepackage{url}

\usepackage{calrsfs}

\usepackage{booktabs}
\newcommand{\ra}[1]{\renewcommand{\arraystretch}{#1}}

\usepackage{soul}
\soulregister\ref{7}
\soulregister\cite{7}
\soulregister\refFig{7}

\usepackage{color}
\definecolor{gray}{rgb}{0.35,0.35,0.35}
\definecolor{blue}{rgb}{0,0,1}
\definecolor{light-blue}{rgb}{0,1,1}
\definecolor{pink}{rgb}{1,0.5,1}
\definecolor{yellow}{rgb}{0.75,0.75,0}
\definecolor{white}{rgb}{1,1,1}
\definecolor{dark-brown}{rgb}{0.2,0.1,0}
\definecolor{green}{rgb}{0,1,0}
\definecolor{dark-blue}{rgb}{0,0,0.8}

\usepackage{epsfig}
\usepackage{epstopdf}
\usepackage{multirow}
\usepackage{array}

\newcommand{\todo}[1]{{\color{red} #1}\normalfont}
\newcommand{\pink}[1]{{\color{pink} #1}\normalfont}
\newcommand{\green}[1]{{\color{green} #1}\normalfont}
\newcommand{\yellow}[1]{{\color{yellow} #1}\normalfont}

\newcommand{\rev}[1]{{\color{black} #1}\normalfont}
\newcommand{\rrev}[1]{{\color{black} #1}\normalfont}

\newcommand\blfootnote[1]{%
	\begingroup
	\renewcommand\thefootnote{}\footnote{#1}
	\addtocounter{footnote}{-1}
	\endgroup
}

\newcommand{\changes}[1]{{\color{black}#1}\normalfont}

\usepackage{xcolor}

\newbox\jsavebox
\newcommand{\jsubfig}[2]{%
	\sbox\jsavebox{#1}%
	\parbox[t]{\wd\jsavebox}{\centering\usebox\jsavebox\\#2}%
	}

\newcommand{\uvec}{{\bf u}} 
 
\newcommand{\vtheta}{{\bf \theta}}

\newcommand{\f}{f_{\vtheta}}

%% file: abstract.tex
Nowadays, as cameras are rapidly adopted in our daily routine, images of documents are becoming both abundant and prevalent.
Unlike natural images that capture physical objects, document-images contain a significant amount of text with critical semantics and complicated layouts.
In this work, we devise a generic unsupervised technique to learn multimodal affinities between textual entities in a document-image, considering their visual style, the content of their underlying text and their geometric context within the image. We then use these learned affinities to automatically cluster the textual entities in the image into different semantic groups. 
The core of our approach is a deep optimization scheme \rev{dedicated for an image provided by the user} that detects and leverages reliable pairwise connections in the multimodal representation of the textual elements in order to properly learn the affinities. 
We show that our technique can operate on highly varying images spanning a wide range of documents and demonstrate its applicability for various editing operations manipulating the content, appearance and geometry of the image.

\ignorethis{
Nowadays, as cameras are rapidly adopted in our daily routine, images of documents are becoming both abundant and prevalent.
Unlike natural images that capture physical objects, document-images contain a significant amount of text with critical semantics and complicated layouts.
Editing the style, location, and content of text within an image can be a tedious and repetitive process. Following color and tonal editing techniques in images, we present a generic multimodal embedding technique for learning the semantic affinities among textual regions in the image, to significantly speed up the editing process. In contrast to previous work, visual features and spatial locality alone cannot fully infer the latent affinities among the many textual elements in the image. 
Therefore, to formulate edit propagation on a document-image, we define an affinity measure among words, which considers their visual style, the content of their underlying text, and their context within the image. 
Our approach is unsupervised, built on an optimization scheme which learns a suitable affinity measure for the multimodal word representations. Our key observation, which motivated the affinity space optimization scheme, is that reliable pairwise connections in the data can guide the network in properly learning the affinities. 
After learning the semantic affinities, the words are clustered in the affinity space, so that all edits performed on a given word are propagated to all words in its cluster. This framework allows the user to perform editing operations on large groups of similar text entities, with just a few clicks.
We show that our technique can operate on highly varying images spanning a wide range of documents and demonstrate its applicability for various editing operations which manipulate the content, appearance and geometry of the image.

Nowadays, as cameras are rapidly adopted in our daily routine, images of documents are becoming both abundant and prevalent.
Unlike natural images that capture physical objects, document-images contain a significant amount of text with critical semantics and complicated layouts.
Editing the style, location, and content of text within an image can be a tedious and repetitive process. Visual features and spatial locality alone cannot fully infer the latent affinities among the many textual elements in the image. 
Following color and tonal editing techniques in images, we present a generic multimodal embedding technique for learning the semantic affinities among textual regions in the image, to significantly speed up the editing process.
To formulate edit propagation on a document-image, we define an affinity measure among words, which considers their \emph{visual} style, the \emph{content} of their underlying text, and their \emph{context} within the image. 
Our approach is unsupervised, built on a novel optimization scheme which learns a weighted non-linear combination of a high dimensional multimodal affinity space. Our key observation, which motivated the affinity space optimization scheme, is that reliable pairwise connections in the data can guide the network in properly weighting the initial features. 
Our framework allows the user to easily perform various types of editing operations on large groups of similar text entities, with just a few clicks.
We show that our technique can operate on highly varying images spanning a wide range of documents and demonstrate its applicability for various editing operations which manipulate the content, appearance and geometry of the image.

}

%% file: acm_classification_terms.tex

\ccsdesc[500]{Computing methodologies~Unsupervised learning}
\ccsdesc[300]{Computing methodologies~Neural networks}
\ccsdesc[300]{Computing methodologies~Learning latent representations}
\ccsdesc[500]{Computing methodologies~Instance-based learning}
\ccsdesc[500]{Computing methodologies~Image manipulation}

%% file: figures/teaser/teaser.tex
\begin{teaserfigure}
\vspace{-8pt}
\centering%
	\jsubfig{\includegraphics[height=6.9cm]{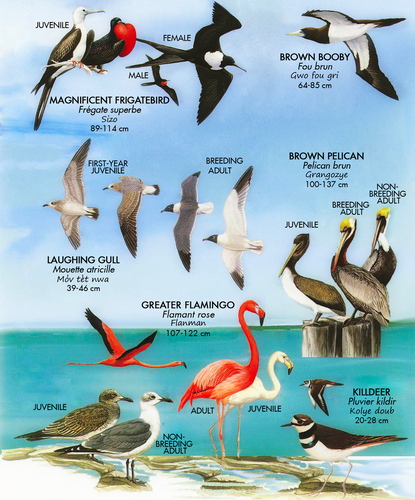}}
	{Input Document-Image}%
	\hfill%
	\jsubfig{\includegraphics[height=6.9cm]{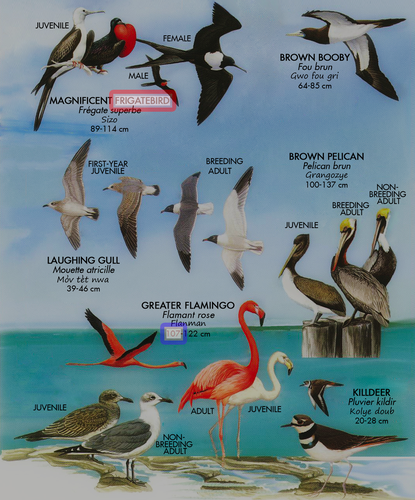}}
	{User Edits}%
	\hfill%
	\jsubfig{\includegraphics[height=6.9cm]{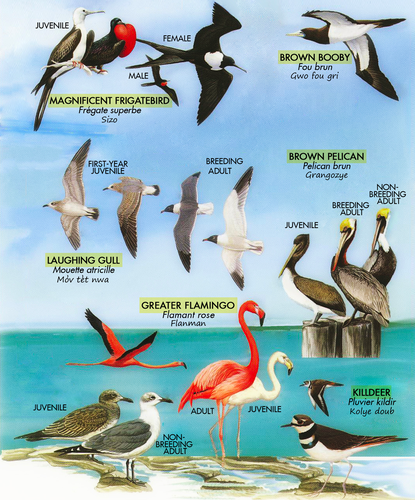}}
	{Edited Document-Image}%
\vspace{-8pt}
\caption{Given a document-image (left), we learn multimodal affinities among the textual entities. The user can then select words to edit. For example, the user can select to highlight the word ``FRIGATEBIRD'' marked in red or delete the word ``107'' marked in blue (center). Our method propagates the editing operations onto words that are semantically and visually similar, highlighting all bird names and removing their size information (right). Image courtesy: Société Audubon Haiti.
}
\label{fig:teaser}
\end{teaserfigure}

%% file: intro.tex
\section{Introduction}

\blfootnote{$^{\dagger}$ Work done while Omri Ben-Eliezer and Hadar Averbuch-Elor were with Amazon Web Services.}

We, as humans, commonly view natural images as those that contain physical objects, such as people, cars or flowers. However, a significant portion of the images uploaded to the internet capture documents.
This calls for automated generic methods to understand, organize and manipulate text in images, and specifically in document-images. A fundamental ability useful for this type of tasks is the automatic \emph{clustering} of text in images, where each cluster corresponds to a unique ``category'' of words, similar in meaning and context within the document. For example, for the \rev{infographic} depicted in Figure \ref{fig:teaser}, possible text categories could be species names, Latin names, species size, gender, or age (juvenile vs. adult). 

\rev{Documents come in large variety. However, while there are countless datasets of images capturing groundable objects which include various granularities such as per-pixel segmentation maps,   
datasets of document-images are sparse and semantically labeling new documents is prohibitively expensive. 
}
Therefore, in this work, we propose a generic unsupervised method to learn meaningful affinities among textual elements in document-images, allowing to cluster them into (implicit) categories, whose number is not given in advance.
Our method processes the image in several steps. First, it obtains text entities from the image using an off-the-shelf OCR technique, and extracts several types of high-dimensional features which represent the \emph{visual} style of these entities, \emph{geometric} context within the document, and  \emph{semantic} content of the underlying words. As it turns out, clustering in this high-dimensional feature space does not yield good results, as the affinities among textual elements are noisy and some of the semantic groups are intermixed. As we show, this also holds true for features originating from a single modality (see, e.g., the style-based features visualized in Figure \ref{fig:no_optimization}). 

Thus, we devise a constraint-based optimization scheme, which generates a low-dimensional representation for the words in the document, especially suitable for clustering purposes (see Figure \ref{fig:ablation_optimization1}). This optimization scheme is our main technical contribution; \rev{for each image provided by the user, it leverages reliable pairwise connections extracted automatically from the initial representation of the textual entities and optimizes the learned affinities.} 
During training, the latent representations of pairs that constitute ``must-link'' constraints are drawn closer together, while ``cannot-link'' pairs are pushed towards far-away parts of the latent space. With the learned affinity, one can group close-by entities in the latent space. Furthermore, if needed, our pair-based approach enables augmenting the information with labeled pairs to constitute a semi-supervised framework, allowing the user to refine the affinity space with several user-drawn scribbles that describe pairs of words that should or should not be grouped together. 
\ignorethis{
To learn the semantic affinities, we take an unsupervised approach. Given an input image, we first extract the words using an off-the-shelf OCR framework. We extend words to \emph{contextual-lines} by grouping horizontally close-by words. To initialize the latent word representation space, we extract multimodal features from the words, obtaining a high-dimensional feature representation.

The key idea in our work is that a \emph{correct} affinity measure for these high-dimensional representations can be learned by leveraging reliable pairwise connections between words. Specifically, we define \emph{must-link} connections between pairs of similar words and \emph{cannot-link} between words whose initial representation differs substantially. 
We then utilize the constraints to learn a mapping of the high-dimensional representations into a latent low-dimensional affinity space: during training, pairs that constitute must-link constraints are drawn closer together, while cannot-link pairs are pushed towards far-away parts of the affinity space. With the learned affinity, one can group close-by entities in the affinity space, and then propagate edits made to a single element to all elements within its group. Furthermore, if needed, our pair-based approach enables augmenting the information with labeled pairs to constitute a semi-supervised framework, allowing the user to refine the affinity space with several user-drawn scribbles that describe pairs of words that should or should not be grouped together. 
}

\rev{
The ability to learn meaningful textual affinities in images can have a multitude of applications, such as document editing, similarity analysis, or locality sensitive hashing of document-images. 
To practically demonstrate one  useful application of our framework, we show how the cluster-based representation can very easily accelerate text editing tasks in images, even when one does not have access to the source code of the document 
(for example, when a graphic designer wishes to generate a slightly modified version of a text layout found on the web).
}

Editing textual content in document-images can be a tedious task, especially if the amount of text entities is large, as the editing operations are repetitive in many cases. Some examples of such potentially repetitive tasks include changing the font type, size, or color for all content of a given type, updating all occurrences of date and time in a schedule sheet, aligning the names of all dishes in a menu to be in the same column, or removing content of a certain type (see Figure \ref{fig:teaser}).
Efficiently propagating a sparse set of user edits in an image has received considerable attention in the context of color and tonal editing \cite{li2008scribbleboost,farbman2010diffusion,chen2012manifold}. Typically, editing operations are propagated to pixels of similar appearance, bypassing the need to perform a careful region selection and matting. 
In our setting, we propagate textual edits in the document-image by simply applying an edit performed on a single text entity to its whole cluster.

\ignorethis{In our work, we are interested in propagating textual edits in document-images.  Unlike the traditional setting where similarity is measured in terms of appearance and position only, textual regions contain latent semantics that must be accounted for in defining their intra-affinity, which visual features and spatial locality alone cannot fully infer. 
Thus, we present a technique that learns the semantic affinities among textual regions, and consequently allows to propagate edit operations among words that are close-by in a latent affinity space. Given the extracted words in the document-image, their affinity space is formed by jointly considering their \emph{visual} style, the semantic \emph{content} of their underlying text, and their \emph{context} within the document.}

\input{figures/ablation/no_optimization.tex}
\input{figures/ablation/ablation_optimization.tex}

We supplement our clustering and edit propagation approach for document-images with a new publicly available dataset of $50$ document-images 
spanning multiple \rev{types of infographics, 
including menus, schedules, brochures, and maps, as well as dense documents} which we quantitatively evaluate our approach on.

 Explicitly stated, our main contributions are:
 \begin{itemize}
 \item A constraint-based optimization scheme for learning affinities among words in document-images, taking into account multimodal features of the words.
 \item A generic unsupervised approach to group textual elements in documents by similarity, without the need to define semantic categories (e.g.~title, paragraph, caption) in advance, useful for textual editing in images. 
 \item A new dataset of document-images of various types, useful as a baseline for future works on document-image analysis.
  \end{itemize}

%% file: figures/ablation/no_optimization.tex
\begin{figure}
		\centering%
    \jsubfig{\fbox{\includegraphics[height=3.00cm]{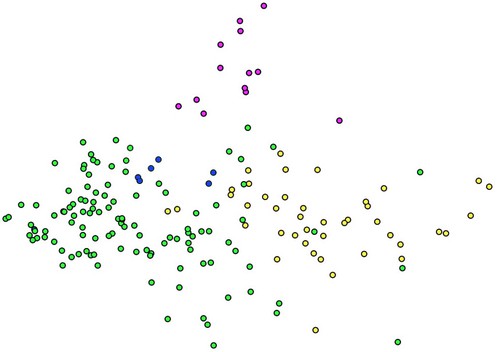}}}
	{}%
    \hfill
	 \jsubfig{\fbox{\includegraphics[height=3.00cm]{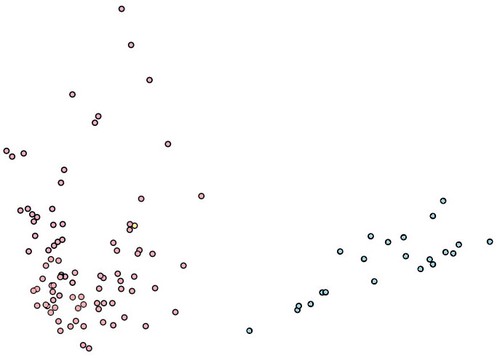}}}
	{}%
\caption{\changes{Style-based affinity space \emph{without} our optimization scheme. Above we visualize the affinity space obtained from style-based features only on the examples illustrated in Figures \ref{fig:ablation_optimization1} (left above) and \ref{fig:overview} (right above). The points are colored according to the clustering provided in the corresponding figures. As the figure illustrates, even for features originating from a single modality (style-based in this case), the initial affinities are noisy and thus challenging to cluster. }
}
\label{fig:no_optimization}

\end{figure}

%% file: figures/ablation/ablation_optimization.tex
\begin{figure}
		\centering%
		\jsubfig{\fbox{\includegraphics[height=3.699cm]{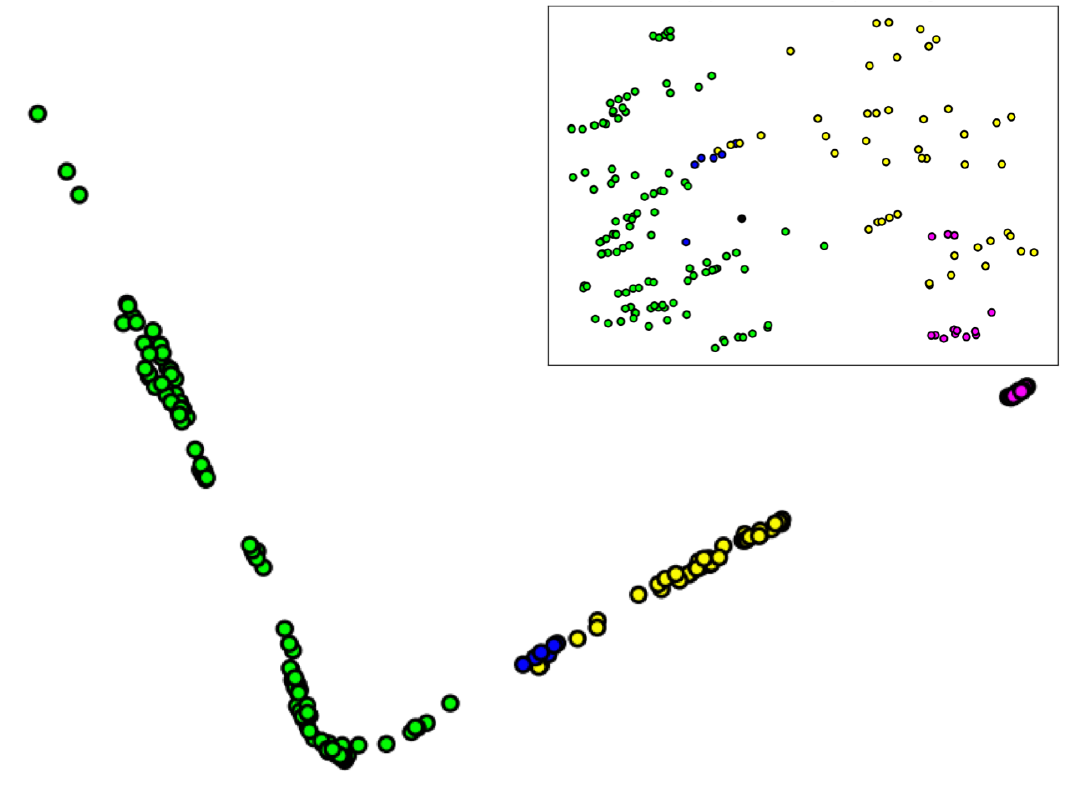}}}
	{}
	\hfill
	\jsubfig{\includegraphics[height=3.75cm]{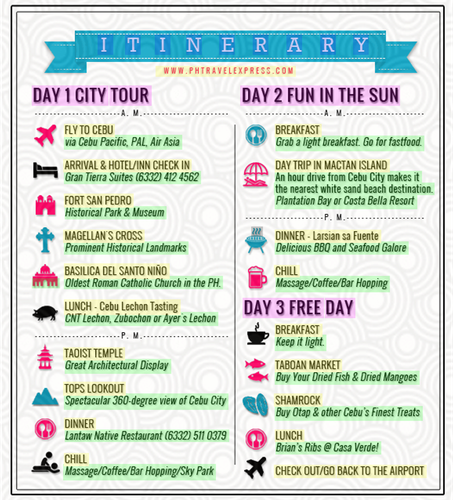}}
	 {}

\caption{\rrev{Above we use PCA to visualize our affinity space before and after our optimization scheme (left). The projected points are colored according to the output grouping result (right). As the figure illustrates, before the optimization stage some of the semantic groups are intermixed and would not allow for measuring high-quality affinities (left image, top-right corner). After the optimization, the affinities are significantly more tightly clustered.} Original image © PH Travel Express.
}
\label{fig:ablation_optimization1}

\end{figure}

%% file: related.tex
\section{Related Work}
In what follows, we elaborate on the most closely related work.

\subsection{Textual Analysis in Images}
Automated structural analysis of document-images has been of interest for more than thirty years \cite{Kasturi2002}. Earlier works relied on predetermined logical rules and heuristics to solve either high level global tasks such as layout and hierarchy analysis, and skew detection, or local tasks such as high-accuracy OCR detection; see e.g. \cite{Akiyama1990AutomatedES, Ogorman1993} and the references within, as well as the comprehensive book of Baird et al.~\shortcite{Baird1992StructuredDI}. However, due to their reliance on heuristics with which it is hard to capture all possible end-cases, these classical approaches typically experience difficulties when performing global analysis of documents with complex layouts. 
Breuel~\shortcite{Breuel2002, Breuel2003HighPD} proposes efficient and exact geometric algorithms for some tasks typically arising in document layout analysis, thus overcoming the need for heuristics in these tasks. 

Recent works have used neural networks to conduct global analysis of document images in a more robust and nuanced manner. Most notably, Yang et al.~\shortcite{yang2017learning} follow a multimodal approach to extract the semantic structure from document images. The structure extraction is viewed as a pixel-wise segmentation task  -- i.e., as the task of segmenting the pixels of the document image into several type of predefined labels, with examples such as ``paragraph'', ``list'', and ``section heading''. To train their network, they devise a suitable synthetic document generation process. Their work shows that joint textual and visual embeddings of pixels in a document image aids in segmentation tasks. While our work also uses a multimodal embedding for the textual elements in the image, it significantly diverges from \cite{yang2017learning} in that it is unsupervised, and performs grouping without the need for pre-defined labels, allowing to flexibly group the words in the document into an unknown number of clusters, whose characteristics are not known in advance.

\rev{Many of the examples of documents we consider in this paper are \emph{infographics}; these are documents which present information or data visually, in a way that is effective and easy to digest for the reader. Unlike dense documents, which typically admit a more standard and unified structure, infographics come in a variety of visual structures and automatically extracting information from them requires specialized machinery. There has been a multitude of works on parsing and understanding infographics. Kembhavi et al.~\shortcite{Kembhavi2016} propose Diagram Parse Graphs, a novel logical representation for diagrams, and use them for question-answering on diagrams; 
Bylinskii et al.~\shortcite{visually1, visually2} extract predictive visual information from infographics by studying connections between the textual and the visual elements;
Poco and Heer~\shortcite{Poco2017} automatically parse and analyze visual encodings in charts and graphs;
Chen et al.~\shortcite{ChenWWWQ20} devise a deep learning based approach for designing timeline infographics. 
Lu et al.~\shortcite{Lu2020} analyze infographics following the Visual Information Flow, the implicit semantic structure in which graphic elements are organized so as to convey information to the reader. Common to all of these works is that they analyze the role of textual elements in infographics as part of the general visual structure. However, none of them directly addresses the task of clustering ``similar'' text entities, discussed in this paper. 
}

Other previously investigated tasks that were shown to benefit from a joint visual and textual representation include, e.g., visual question answering \cite{VQA, Goyal2018, vqa-cp} and image captioning \cite{Donahue17, Karpathy17}.
Detection and understanding of text in the wild -- recognizing text in natural scene images -- has also been thoroughly investigated, see e.g. \cite{Jaderberg2016}.

\subsection{Image Editing and Edit Propagation}
Image editing is a central application in computer graphics and has significantly advanced over the past several decades. 
One of the first works in automatic image editing \cite{perez2003poisson} presents a method to seamlessly edit regions in images, based on Poisson equations.
Patch-based approaches have also shown to be useful throughout the years (e.g., \cite{barnes2009patchmatch}).
In recent years, most of the computer graphics image editing techniques have been based on deep neural networks.
Specifically, various works have shown that training generative models using a GAN framework \cite{goodfellow2014generative} is effective for image generation and manipulation.
As an example, \cite{liu2018image} recently achieved remarkable results for image completion.

Our work adds to the long line of works that develop automated editing procedures aiming at reducing human effort within the editing process. Specifically, there have been various works on propagating visual image editing. For example, Li et al.~\shortcite{li2008scribbleboost} show that the combination of edge-aware interpolations with a classification step is effective for speeding up editing of image regions via user scribbles. An and Pellacini \shortcite{an2008appprop} show that edit propagation can be viewed as a linear problem, where a sparse set of user edits combined with structural observations on the underlying linear system suffice to robustly propagate edits between different entities. Xu et al.~\shortcite{xu2009efficient} show that adaptive clustering with the help of $k$-d trees can significantly speed up the approach of \cite{an2008appprop} without affecting the visual quality of the resulting edits. Farbman et al.~\shortcite{farbman2010diffusion} observe that, for the sake of evaluating the similarity between pixels for editing purposes, it is helpful to replace the usual Euclidean distance between pixels with so-called diffusion distances. Chen et al.~\shortcite{chen2012manifold} consider a different approach for edit propagation, which seeks to maintain the topology of the image through preservation of its manifold structure. Xing et al.~\shortcite{AutocompletePainting} devise a system that automatically completes repeated patterns while painting, allowing the user to avoid repetitive drawing tasks. Finally, a recent work of Meyer et al.~\shortcite{DeepVideoPropagation} presents an intriguing deep learning framework to propagate color changes along videos, combining local and global approaches to ensure the stability and consistency of the solution.

\subsection{Document Editing} While we mainly target document-images and are only provided with an image of a document, our approach can also be used to analyze documents in their raw format. Commercial software products for creating and managing documents with highly complex layouts include, e.g., Adobe InDesign\footnote{https://www.adobe.com/products/indesign.html} and FrameMaker\footnote{https://www.adobe.com/products/framemaker.html}, and Microsoft Publisher\footnote{https://products.office.com/en-us/publisher}, among many others.  
\cite{Saund2003} propose a content-aware toolkit to perform various editing operations on documents, relying on local grouping of the text elements. 
To this end, Project Naptha\footnote{https://projectnaptha.com/} is a popular OCR-based Google Chrome extension allowing the user to perform multiple basic editing operations in web-pages, including copying, highlighting, editing and translation of text.
While the above systems and works provide numerous sophisticated tools for high-quality editing of documents with complex layouts, we are not aware of any product and/or previous work that allows automatic global similarity analysis and grouping as is done in our work.
Incorporating our global similarity analysis capabilities into existing solutions for document analysis might serve as an interesting venue for future research, making these solutions more similar to the way humans perceive documents.

\input{figures/overview/overview_fig.tex}

\subsection{Affinities and Clustering in Images}

Our approach aims to learn an affinity, i.e., a measure of similarity, between (textual) objects in (document-)images. Analysis of similarity in images has been a widely studied task. For example, ~\cite{garces2014} devise a style-based similarity measure for objects in clip art images, relying only on visual features (and not on content): specifically, on color, shading, texture, and stroke features. 

At the core of our approach is a deep optimization scheme for clustering textual data into an unknown number of clusters.
Data clustering is one of the most fundamental problems in data analysis, highly applicable to different fields of science, and is especially essential in an unsupervised learning scenario. 
Traditionally, clustering techniques focused on the grouping task and assumed a representation is provided as input. However, recently, clustering methods have coupled the feature learning and the grouping stages into one unified deep optimization scheme. 
Xie et al. \shortcite{xie2016unsupervised} use a stacked autoencoder to learn a deep representation which is easier to group onto a known number of cluster centroids. Later works (e.g., \cite{yang2016joint,dizaji2017deep}) have been able to boost performance for images using convolutional networks. More closely related to our work, Shah and Koltun \shortcite{Shah2018} and Fogel et al.~ \shortcite{fogel2018clustering} use pairwise constraints to learn a cluster-friendly representation without prior knowledge on the number of clusters in the data. 
These works operate on the images directly, leveraging a reconstruction loss to learn an initial embedding space. In our work, we are interested in learning semantic affinities between words in the image and thus initialize the latent space using multimodal features that encode both the style and the content of the extracted words.

%% file: figures/overview/overview_fig.tex
\begin{figure*}
	\centering%
    	\jsubfig{\includegraphics[height=5.42cm]{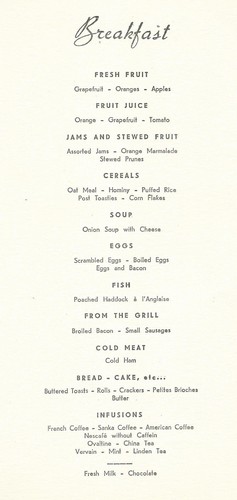}}
	{Input}%
	\hfill%
	\jsubfig{
	\includegraphics[height=5.42cm]{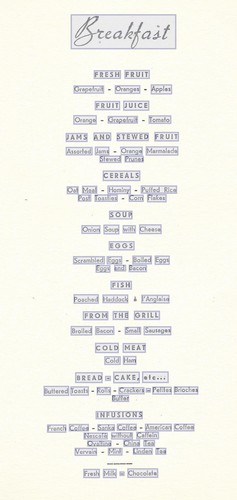}	
	\includegraphics[height=5.42cm]{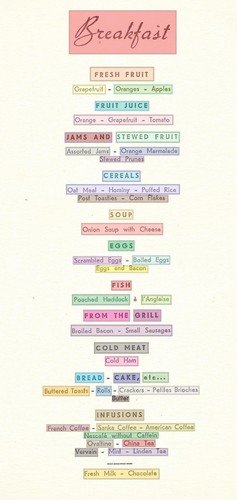}
		}
	{Identifying basic elements}%
	\hfill%
	\jsubfig{
	\includegraphics[height=5.42cm]{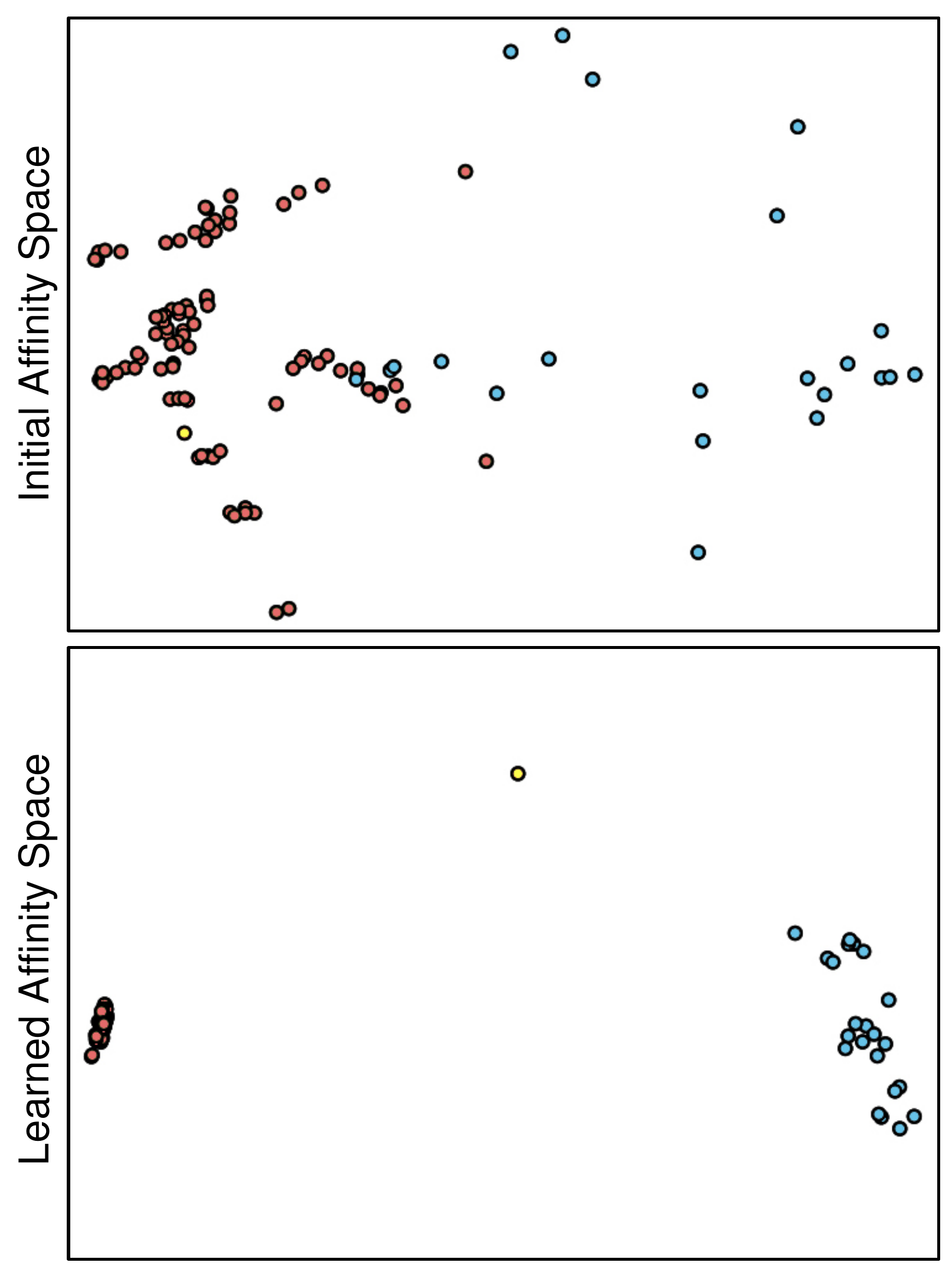}
			\includegraphics[height=5.42cm]{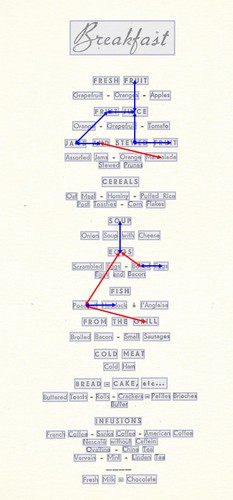}
			}
	{Constraint-based affinity learning}%
	\hfill%
	\jsubfig{
	\includegraphics[height=5.43cm]{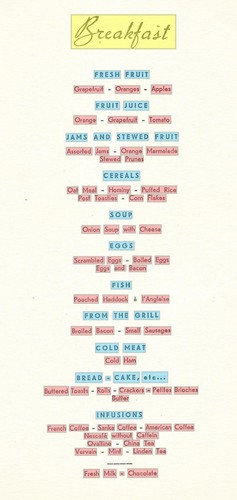}	}
	{Output}%
	
\caption{Given a document-image (left), we first extract words and contextual-lines within the image. To initialize the latent affinity space, we extract multimodal features for each word in the image. We then learn the semantic affinities among words in the image by leveraging reliable pairwise connections in the form of must-link (marked with blue arrows) and cannot-link (marked with red arrows) constraints. Finally, we use the learned affinities to obtain semantic groups for edit propagation tasks (right). Image courtesy: Compagnie générale transatlantique.
}
\label{fig:overview}

\end{figure*}

%% file: overview.tex
\section{Overview}

In this work, we consider the problem of learning affinities for clustering textual elements in an image, and demonstrate the use of this clustering primitive for edit propagation purposes.
\rev{
Our underlying assumption is that we do not have access to any meta-data on the analyzed document, and so our learned affinities and resulting clustering are based purely on the information extracted from the image of the document. The existence of such meta-data, however, would significantly ease the clustering task as it would provide additional constraints which can be leveraged for learning the textual affinities.
}

Our solution first extracts multimodal features for any textual element in the image; then, it learns a suitable notion of affinity for the elements within the document in an unsupervised manner; and finally, the affinity notion is leveraged to cluster words into the desired groups. In more detail, our solution constitutes of the following steps.

\subsubsection*{Identifying basic elements} 
The main challenge in our context is the understanding of text semantics. Thus, while the basic elements in previous works have usually been pixels, here we use words as the basic elements. As OCR-related works have essentially solved the challenge of word-level identification, we first use an off-the-shelf OCR method to extract words and other symbols in the image, and use these as our basic textual elements. Then, as basic elements located very close to each other tend to be semantically related, we group
our basic elements into \emph{contextual-lines} (see Section \ref{sec:units}).

\subsubsection*{Extracting multimodal features}
To be able to devise an affinity measure for the basic elements in our context, we need a suitable representation that will capture their ``essence''. The representation, described in detail in Section \ref{sec:features}, captures multimodal features of the textual elements, which can be divided into three families:
\begin{itemize}
    \item Style-based features, which refer to the visual appearance of words. Using a pre-trained classifier, we extract visual features related to font characteristics for each word.
    \item Content-based features of words capturing the semantic meaning of the word in a broader context, considering its enclosing contextual-line.
    \item Geometric features, concerned with the size and location of the words within the document.
\end{itemize}

\subsubsection*{Constraint-based affinity learning}
Our next main challenge is to learn an affinity measure for our multimodal space. 

\paragraph{Pairwise constraints}
As our approach is unsupervised, to guide the learning we extract \emph{pairwise constraints}, which are reliable pairwise connections between words in the document. We derive several types of pairwise connections in two categories: \emph{must-link} constraints that correspond to pairs that are very similar according to our analysis (the training process will pull such pairs towards each other in the affinity space), and \emph{cannot-link} constraints, corresponding to pairs of ``very different'' words that should be far apart in the affinity space (see Section \ref{sec:constraints}).

\paragraph{Affinity Space Optimization} 
Our objective now is to learn an affinity measure. Formally, suppose that the multimodal representations obtained above are given as $D$-dimensional vectors. We consider a low-dimensional latent space $\mathbb{R}^d$ coupled with an affinity measure $\sigma$ for vectors in $\mathbb{R}^d$, which outputs $1$ for ``similar'' vectors and close-to-0 values for ''very different'' vectors. In practice, $d \ll D$.
Our objective is to train a network $\f \colon \mathbb{R}^D \to \mathbb{R}^d$
with the goal of having $\sigma(\f(Z_i), \f(Z_j)) \approx 1$ (high affinity value) for pairs of similar words with representations $Z_i, Z_j$, and a much smaller affinity value when $Z_i$ and $Z_j$ are not similar.
For training, we use a Siamese architecture, feeding pairs of must-link/cannot-link words to two identical copies of the network along the training process; roughly speaking, the loss we aim to minimize during training ``punishes'' us for must-link constraint pairs $Z_i, Z_j$ with $\sigma(\f(Z_i), \f(Z_j)) \ll 1$ and for cannot-link pairs when $\sigma(\f(Z_i), \f(Z_j)) \approx 1$. See Section \ref{sec:optimization} for more details (and Section \ref{sec:implementation} for further implementation details).

\subsubsection*{Clustering in affinity space}
Finally, we use our now trained network $\f$ and the learned affinities in order to obtain the desired groups for edit propagation (see Section \ref{sec:clustering}).

%% file: words.tex
\section{Basic Textual and Contextual Elements}
\label{sec:units}
We first extract the basic textual elements 
in the input image. To do so, we use standard techniques to detect word bounding boxes and recognize the words within these bounding boxes \cite{smith2007overview}. Isolated words are not sufficient to effectively analyze natural language. Thus, we augment our basic textual elements with a contextual notion by grouping close-by words and associating a \emph{contextual-line} to each extracted word.

We follow classical approaches in the literature  (see e.g. \cite{Baird1992StructuredDI}) and pick the contextual-lines as the connected components of a suitable sparse graph representation of the document-image.
We obtain contextual-lines by grouping horizontally adjacent words that contain similar geometric attributes. Specifically, we construct an edge-weighted graph of word boxes, where the weight between a pair of words is the horizontal space-to-height ratio between them, normalized by the aspect ratio (image width divided by image height). Formally, given two word bounding boxes $B_i$ and $B_j$ whose $y$-coordinates have a sufficiently large overlap, denote the heights of the boxes $B_i, B_j$ by $h_i, h_j$ respectively, the horizontal space between them by $s$, and the aspect ratio of the image by $r$. We then define the weight of the corresponding edge by 
$$ w(B_i, B_j) = \frac{s \cdot r}{\max(h_i, h_j)}.
$$
We then define a sparse graph representation by removing edges whose weights are above a predefined threshold (empirically set to 0.1). The connected components in this sparse graph form our contextual-lines.

It should be noted that we deliberately avoid more sophisticated approaches to extract lines and paragraphs (e.g.,~\cite{Breuel2002, Breuel2003HighPD}). 
Indeed, we seek to group words into contextual-lines as conservatively as possible, so as to avoid grouping words with different contexts into the same contextual-line at all costs.
We then use a neural network, described in the next sections, to learn the non-local affinities. 

%% file: features.tex
\input{figures/font/font.tex}

\section{Affinity Space Initialization}
\label{sec:features}
As described in the previous section, the basic textual element we consider is a word unit, henceforth denoted by $U_w$, which can be described visually by its associated image region or semantically by its underlying content. To augment words with a contextual notion, each word is associated with a contextual-line, henceforth denoted by $U_l$. 
We initialize our affinity space by extracting multimodal features for each word in the image and concatenating them to obtain our aggregated word representation $Z(U_w,U_l)$. Explicitly stated, our initial representation $Z$ is defined as  
\begin{align*}
    Z(U_w,U_l) = \large[ \textbf{f}_s(U_w),  
     \textbf{f}_c(U_w, U_l), \textbf{f}_g(U_w) &\large],
\end{align*}
where $\textbf{f}_s(U_w)$ encodes the visual style, $\textbf{f}_c(U_w)$ encodes the semantic content and $\textbf{f}_g(U_w)$ encodes geometric attributes. Next, we describe these multimodal encoders in more detail.

\subsection{Style Encoder} 
\label{sec:style}
The style encoder $\textbf{f}_s(U_w)$ aims to extract the visual appearance of a word $U_w$. In our setting, visual appearance can roughly be estimated using font characteristics. Therefore, we pre-train a classifier to distinguish between multiple font characteristics such as font-family, font-weight (regular/bold) and style (regular/italics). Specifically, we employ a Residual Network (ResNet) architecture \cite{he2016deep} as a visual feature extractor trained on a synthetic dataset we generate using an open source synthetic data generator\footnote{https://github.com/Belval/TextRecognitionDataGenerator}. 

Figure \ref{fig:font} illustrates samples belonging to three different classes (out of the 138 classes used during training). As illustrated in the figure, a unique configuration of font-family, font-weight and style is determined for each class. \rev{The content, background, color, and font-size vary for each sample, to encourage invariance to these properties. Note that while the text color could be considered part of its \emph{style}, we noticed that in many cases the color may vary (even within the same contextual element), making it an unreliable cue. }

At test-time we use the last layer of the trained classifier (that is, the input to the last fully connected layer in the ResNet architecture) as a visual feature representation for each basic unit in the document-image.

\subsection{Content Encoder}
\label{sec:content}
The content encoder $\textbf{f}_c(U_w, U_l)$, captures semantic language features of a word $U_w$. Contextualized language-model encoders are recently gaining increasing popularity, as they capture the semantic meaning of the word in a broader context, considering its enclosing phrase. Thus, we employ a contextualized language-model encoder which operates on a word $U_w$ and its enclosing contextual-line $U_l$, thereby capturing the word meaning in the context of that phrase. 

In our work, the content encoder $\textbf{f}_c(U_w, U_l)$ is a concatenation of the state-of-the-art forward and backward models detailed in Akbik et al.~\shortcite{akbik2018coling}. The implementation we used was pre-trained on a large corpora of unlabeled data, and no additional tuning was performed. Please refer to \cite{akbik2018coling} for further details on the model architecture and training.

\subsection{Geometry Encoder}
\label{sec:geometry}
Lastly, as document entities tend to follow aligned patterns, we employ a simple descriptor to capture geometric attributes. Specifically, for each word $U_w$ with associated image coordinates (normalized to $ \left[0,1 \right]$), the geometry encoder
$\textbf{f}_g(U_w)$ is defined as follows:
\begin{equation}
    \textbf{f}_g(U_w) = \left[ x, y, w, h \right],
\end{equation}
where $\left(x,y\right)$ is the top-left corner of the bounding box of $U$, and $w$ and $h$ are the width and height of the bounding box, respectively.

%% file: figures/font/font.tex
\begin{figure}

	\centering%
    	\jsubfig{\includegraphics[height=1.61cm]{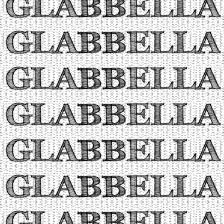}}
	{}%
	\hfill%
	\jsubfig{
	\includegraphics[height=1.61cm]{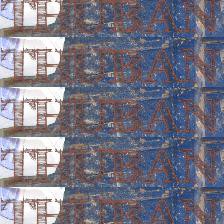}	}
	{}%
	\hfill%
		\centering%
    	\jsubfig{\includegraphics[height=1.61cm]{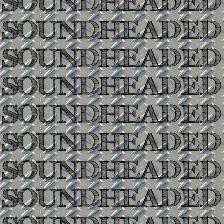}}
	{}%
	\hfill%
	\jsubfig{
	\includegraphics[height=1.61cm]{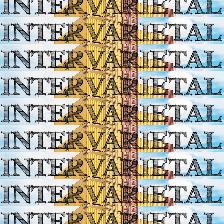}	}
	{}%
	\hfill%
	\jsubfig{
	\includegraphics[height=1.61cm]{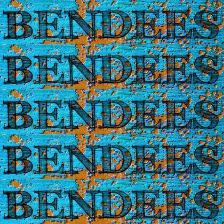}	}
	{}%
	\\
	\vspace{1pt}
    	\jsubfig{\includegraphics[height=1.61cm]{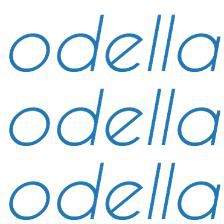}}
	{}%
	\hfill%
	\jsubfig{
	\includegraphics[height=1.61cm]{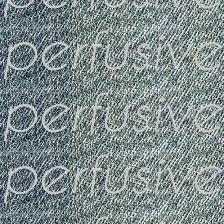}	}
	{}%
	\hfill%
		\centering%
    	\jsubfig{\includegraphics[height=1.61cm]{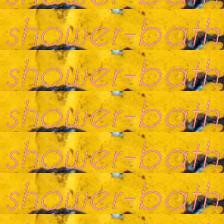}}
	{}%
	\hfill%
	\jsubfig{
	\includegraphics[height=1.61cm]{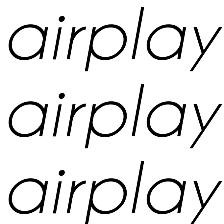}	}
	{}%
	\hfill%
	\jsubfig{
	\includegraphics[height=1.61cm]{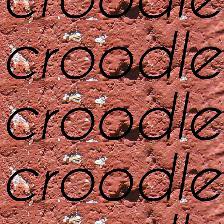}	}
	{}%
	\\
		\vspace{1pt}
		\centering%
    	\jsubfig{\includegraphics[height=1.61cm]{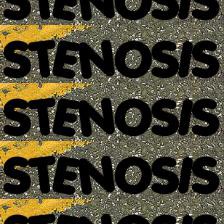}}
	{}%
	\hfill%
	\jsubfig{
	\includegraphics[height=1.61cm]{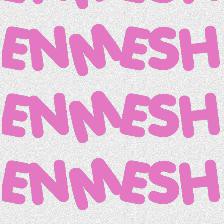}	}
	{}%
	\hfill%
		\centering%
    	\jsubfig{\includegraphics[height=1.61cm]{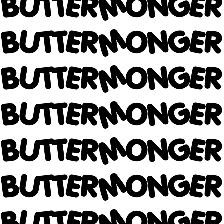}}
	{}%
	\hfill%
	\jsubfig{
	\includegraphics[height=1.61cm]{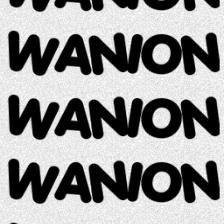}	}
	{}%
	\hfill%
	\jsubfig{
	\includegraphics[height=1.61cm]{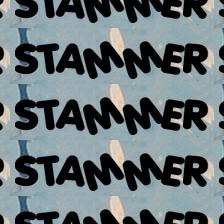}	}
	{}%

\caption{Training samples used for training a style encoder. In each row, we randomly select samples belonging to the same class. As illustrated above, the classes vary in font characteristics, including font-family, font-weight and style.  
}
\label{fig:font}

\end{figure}

%% file: optimization.tex
\input{figures/constraints/constraints.tex}

\section{Constraint-based Learning of Affinities}
\label{sec:optimization_full}
Our approach crucially relies on \emph{learning} an appropriate \emph{affinity} measure between the multimodal representations of words in the document.
In this section we describe how to learn such a measure. The goal is that pairs of words with high similarity in terms of style, content, and geometry will be close by according to the learned affinity, whereas pairs of words that are not similar will be far apart in this affinity space.

Our setting is unsupervised; to learn a suitable affinity measure, we must be able to automatically generate constraints, which will then serve as a basis for the learning process.
To this end, we generate two types of pairwise constraints (i.e., constraints between pairs of words) in which we are relatively confident: \emph{must-link} constraints, indicating that a pair of words are highly similar, and should have a small distance between them; and \emph{cannot-link} constraints, for pairs of words that are ``very different'', implying that the distance between them should be high. 
The constraints are described in-depth below, in Subsection \ref{sec:constraints}; see Figure \ref{fig:constraints} for a visualization. 
They are then fed to a Siamese network, which outputs an affinity measure (a real number between zero and one) given any pair of word units, see Subsection \ref{sec:optimization}.

\subsection{Generating Pairwise Constraints}
\label{sec:constraints}
We employ three different types of constraints: intra-contextual must-link constraints of words that reside within the same contextual-line, inter-contextual must-link constraints and inter-contextual cannot-link constraints. To generate inter-contextual constraints, we examine high-confidence pairs using a set of rules combining the different modalities \rev{and consolidate the results over the multiple domains (i.e. by taking constraints which do not violate any of the modalities).} Finally, we balance the set of constraints by downsampling the constraints such that must-link constraints constitute $60\%$ of the total number of constraints.
Next, we describe our intra- and inter- contextual constraints.

\subsubsection*{Intra-Contextual Constraints}
In Section \ref{sec:units}, we defined the concept of a contextual-line as a basic unit that captures the context of a word. We conservatively grouped only horizontally close-by words with similar geometric properties. Thus, to encourage words within the same contextual-line to output a similar representation, we define intra-contextual must-link constraints, which connect pairs of words belonging to the same contextual-line. 

\subsubsection*{Inter-Contextual Constraints}
In addition to the local intra-contextual must-link constraints, we would like to encourage high affinity pairs, which are not necessarily spatially close, to be pulled closer together. Likewise, to avoid the trivial solution that would collapse all words to a single point, we would like to encourage words that are clearly different to be pushed further apart.

Following the human understanding of what is perceived as close on a document, we define a set of rules spanning the multiple modalities we consider. Inter-contextual \emph{must-link} constraints are defined conservatively between pairs of words which are considered similar in all aspects. In contrast, inter-contextual \emph{cannot-link} constraints are defined between pairs of words that are different in at least one aspect, as a significant difference in even one criterion suggests that these words should not be close in the learned affinity space. To this end, we consider the following criteria:

\paragraph{Visual-style similarity} To extract pairs that are highly similar (or different) in style, we extract pairwise similarities between all visual feature representations, which are described in Section \ref{sec:style}. Pairs of points are considered visually-close if they are connected on a mutual-kNN graph. In other words, two points are visually-close only if they are both among each other's k-nearest neighbors. Likewise, pairs of points are considered visually-different if they are mutually-far. In both cases, we empirically set $k=6$. 

\paragraph{Language-syntax criterion} Words are sorted into three bins according to the types of characters that compose them: entirely uppercase, entirely lowercase, and mixed. Words belonging to the same bin are considered similar in syntax and words belonging to different bins are considered different in syntax.

\paragraph{Language-semantics criterion}
\label{sec:ner_tagging} Named-entity recognition (NER) is a common task in language processing. In this task, words representing semantic entities may be tagged according to some pre-defined categories such as ordinals, percentages, time-expressions, person-names, organizations, and so on.
To encourage semantically based constraints, we employ a NER tagger over word tokens per contextual line. Then, we carefully look at the tagged words and characterize their respective contextual lines, that is, according to the named entities contained within them. Contextual lines defined by multiple types of named entities are considered semantically "noisy" and therefore their words do not yield constraints.
Instead, we generate constraints only for words that characterize their associated contextual-line. Standalone words which are tagged similarly are considered similar in semantics, while standalone words that are tagged differently are considered different in semantics. We use the NER implementation provided by Akbik et al~.\shortcite{akbik2018coling}.

\paragraph{Geometric considerations} We use the bounding-box height dimension to deduce whether two words are similar or not in terms of geometry. Denote the respective bounding-box heights as $h_1$ and $h_2$. Assuming $h_1 > h_2$, we consider the bounding-box height ratio $h_1 / h_2 $. If the ratio is below a predetermined threshold ($1.25$ in our implementation), the words are considered similar in terms of size. Otherwise, they are considered different. 

\subsection{Affinity Space Optimization}
\label{sec:optimization}
Next, we wish to use the constraints to learn a continuous affinity measure for words. \rev{
We solve an optimization problem over \emph{soft} (possibly contradicting) constraints
by employing a Siamese network, that learns a suitable \emph{embedding} of the words into a latent low-dimensional affinity space; See, for example, the visualized affinity space in Figure \ref{fig:ablation_optimization1}.}
The value outputted by the network given a pair of words will then serve as our affinity measure between these words.

Formally, let $\{U_w^1, \ldots, U_w^N\}$ denote the word units in the document-image, and for each word $U_w^i$, let $Z_i$ denote its aggregated representation as defined in Section \ref{sec:features}. We view these representations as $D$-dimensional real vectors, i.e., $Z_i \in \mathbb{R}^D$, where the dimensionality $D$ is determined by the aggregated representations. The latent low-dimensional affinity space is defined as $(\mathbb{R}^d, \sigma)$, i.e., its dimensionality is set to $d$. The value of $d$ is chosen to address the following two issues: On one hand, it should be low enough to avoid the inherent complexities of clustering in high dimensions as much as possible (the so-called ``curse of dimensionality''). On the other hand, recall that our end-goal is to perform a clustering on the latent affinity space, so that each cluster represents a group of similar elements.
Conceptually speaking, each pair of clusters might have a different semantic ``reason'' as to why these clusters are different from each other. Thus, to be able to separate each pair of clusters in the latent affinity space, it makes sense that this space will have roughly one coordinate for each pair of clusters. Therefore, to properly handle document-images where the number of clusters is $k$, our latent dimensionality should be at least $d \approx \binom{k}{2}$.

The affinity function $\sigma \colon \mathbb{R}^d \times \mathbb{R}^d \to [0,1]$ is defined, given $\uvec, \uvec' \in \mathbb{R}^d$, by
\begin{align}
\label{eq:likelihood}
    \sigma(\uvec, \uvec') = \exp \big( -\lVert \uvec - \uvec' \rVert_2^2 \big), 
\end{align}
approximating the similarity function suggested by \cite{Fathi2017SemanticIS}. To this end, a larger affinity value between a pair of words corresponds to ``higher similarity'' between these words. 

Our objective is now to optimize the parameters of a network $ \f: \mathbb{R}^D \rightarrow \mathbb{R}^d$, so that given two word units $U_w^i$ and $U_w^j$ whose corresponding representations are $Z_i$ and $Z_j$, the affinity between the embeddings, which equals
\begin{align}
\label{eq:affinity-final}
\sigma\left(\f(Z_i), \f(Z_j)\right),
\end{align}
would be a good estimate for the actual similarity between these two words. 
We use a Siamese architecture to train the network (see Section \ref{sec:implementation} for implementation details). We pick pairs of words $Z_i, Z_j$ satisfying one of the constraints defined in the previous subsection, and feed them to two identical copies of the network.
We use the function $\sigma$ for its probabilistic interpretation, as $\sigma(\f(Z_i), \f(Z_j))$ in a sense predicts the likelihood that the two points $(U_W^i, U_W^j)$ belong to the same ``category'' in the data. For a pair of points  whose embeddings $\f(Z_i), \f(Z_j)$ are almost identical, we have $\sigma(\f(Z_i), \f(Z_j)) \approx 1$. On the other hand, when the embeddings are entirely different, $\sigma(\f(Z_i), \f(Z_j)) \approx 0$. This function is also numerically stable for training using a cross-entropy classification loss.
Explicitly stated, we train our Siamese network to minimize the following loss:
\begin{align*}
\label{eq:loss}
    L = -& \sum_{\left( i,j \right) \in E_{\text{must-link}} } \,\ \log  \left( \sigma \left( \f(Z_i), \f(Z_j) \right) \right)\\
    -& \sum_{\left( i, j \right) \in E_{\text{cannot-link}} } \log  \left( 1 - \sigma \left( \f(Z_i), \f(Z_j) \right) \right),
\end{align*}

where $E_{\text{must-link}}$ is the set of all pairs $(i,j)$ for which a must-link constraint was added for the pair $(U_w^i, U_w^j)$, and $E_{\text{cannot-link}}$ is defined similarly with respect to cannot-link constraints.

%% file: figures/constraints/constraints.tex
\begin{figure*}
	\centering%
    	\jsubfig{\fbox{\includegraphics[height=6.82cm]{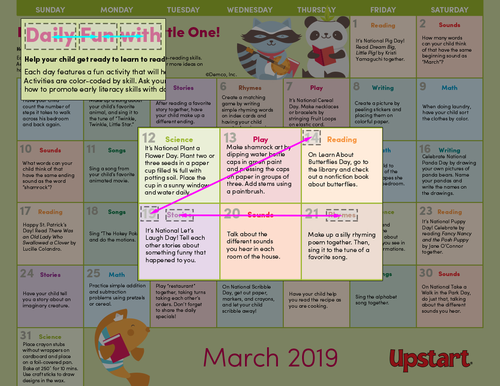}}}
	{Must-link Constraints}%
 	\hfill
		\centering%
    	\jsubfig{\includegraphics[height=6.82cm]{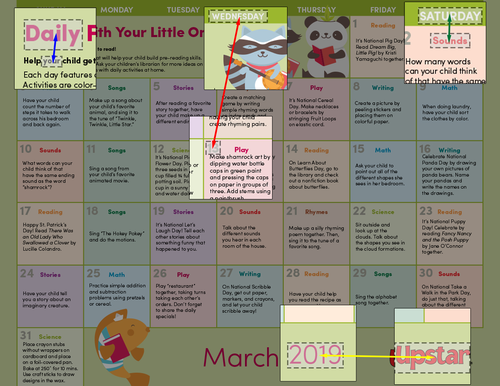}}
	{Cannot-link Constraints}%

\caption{Generating intra- and inter- contextual constraints. Above we illustrate the various constraint types in our framework. Left: We consider must-link constraints between words which either share the same context (marked with light blue arrows) or are highly similar in terms of all the criteria detailed in Section \ref{sec:constraints} (marked with pink arrows). Right: Cannot-link constraints are defined between pairs of words that are different according to visual-style (marked with a yellow arrow), language-syntax (marked with a green arrow), language-semantics (marked with a red arrow) or geometry (marked with a blue arrow). Original image © Demco, Inc.
}
\label{fig:constraints}

\end{figure*}

%% file: clustering.tex
\section{Affinity Space Clustering}
\label{sec:clustering}

In order to demonstrate how the learned affinities can be used, in this work we are interested in propagating textual edits. 
Edit propagation frameworks typically aim at propagating color and tonal edits according to the user scribbles. Thus, previous works use the affinities between pairs of pixels to output the blending weights associated with the different user edits. Unlike the continuous color and tonal edits, in our setting, we seek a binary decision per word. For each word the user selects to edit, we propagate the edit onto all the words that are highly ``similar'' to it, where similarity can be measured in the affinity space described in previous sections.  

Thus, we wish to group the words into several clusters of highly similar elements; this would allow us to simultaneously perform editing operations on all words in a given cluster. To obtain such clusters, we group the contextual-lines according to a mean likelihood estimate that the words within these contextual-lines should be grouped together, where the likelihood estimate corresponds to the notion of affinity obtained in the previous section. Explicitly stated, for a pair of contextual-lines $U_l$ and $U'_l$, we compute the likelihood that $U_l$ and $U'_l$ belong in the same cluster in the data by leveraging a voting mechanism. Specifically, consider the trained network $\f \colon \mathbb{R}^D \to \mathbb{R}^d$ and the affinity function $\sigma$ defined in Equation \eqref{eq:likelihood}. Given each pair of word units $(U^i_w, U^j_w)$ with multimodal represenetations $Z_i, Z_j$, where $U^i_w$ belongs to $U_l$ and $U^j_w$ belongs to $U'_l$, we estimate the affinity of $Z_i$ and $Z_j$ using Equation \eqref{eq:affinity-final}, and then average all such estimates to obtain a likelihood that the two contextual-lines $U_l$ and $U'_l$ should be grouped together.

We then construct an edge-weighted graph of contextual-lines, where the edge connecting a pair of contextual-lines is weighted according to the likelihood estimates. The graph is further pruned by disregarding edges between incompatible contextual-lines (whose bounding box heights are above a constant ratio of $1.25$) and whose weights correspond to a likelihood below a predetermined threshold of $0.75$. 
The words within the connected components of our contextual-lines graph form our final clusters.

%% file: semisupervised.tex
\section{Extension to An Interactive Framework}

Our optimization scheme can naturally be extended to an interactive framework, allowing one to incorporate user inputs in order to refine the affinity space, if needed.
To do so, we first run our algorithm offline, and proceed to augment our automatically obtained pairwise constraints with user-supplied ones. 

Our interactive tool supports a lasso selection tool. In the case of must-link constraints, pairwise constraints are generated among all words in the selection group. For cannot-link constraints, we offer a similar mechanism requiring two clicks for selecting the disjoint groups. We follow up with cannot-link constraints between all pairwise elements from both groups. We keep all pairwise constraints and do not sub-sample the user constraints. Furthermore, we remove automatically-generated constraints which violate the user-specified ones. Please refer to our accompanying video for a demonstration of our interactive system.

%% file: results.tex
\input{figures/architecture/architecture.tex}

\section{Implementation Details}
\label{sec:implementation}
To learn the semantic affinities among words in an image, we use a multimodal encoder neural-network (see Figure \ref{fig:architecture} for a visualization of the network architecture). 
Throughout the evaluation of our method in the experiments described below, the method was run as follows:
During train time, we employ the network as a Siamese encoder with shared weights for each pair of words in the image. To tune the network, we train it using artificial labels originating from our own generated auto-constraints ($y_{c}=1$ for constraint $c \in E_{\text{must-link}}$, and similarly $y_{c}=0$ for constraint $c \in E_{\text{cannot-link}}$) as detailed in section \ref{sec:optimization}. For loss function, we employ cross-entropy loss over the affinity function $\sigma$, also described in the aforementioned section. In addition we perform clipping for gradients of magnitude larger than 5, to increase training robustness.

Our network architecture draws inspiration from \cite{xie2016unsupervised} with slight pruning to the depth of the network, to ensure faster training. Specifically, let $D$ represent the dimensionality of $Z(U_w,U_l)$. Thus the network dimensions are $D-50-2000-d$, where $D$ is determined by the dimensionality of the style, content and geometry encoders, typically set to 512, 4096 and 4 respectively. In our work we set $d=20$, which is low enough to avoid intricacies involved with clustering of high-dimensional data, yet high enough to contain information required for proper separation to clusters on typical documents.

\input{tables/user_study}

For the initialization of the network, we follow the practice of \cite{xie2016unsupervised} and draw random weights for a zero-mean normal distribution, with standard deviation of 0.01. For training, we freeze the embedding layers weights, and train the network for 100 epochs, with learning rate of $1e^{-4}$ and batch size of 32, using an Adam Optimizer \cite{Kingma2015AdamAM}. 
For all our encoder blocks, we use a dropout of 0.2. Furthermore, to speed up the training on denser documents, we allow a maximum of 1000 must-link constraints, randomly-sampling the constraints if needed. Doing so allows us to cap the training runtime at up to one minute on a machine equipped with a single Tesla V100-SXM2 GPU. 

Our method is implemented in Pytorch, and can be further optimized. Our experiments show that the entire pipeline converges within 35 seconds on average documents of 200 words, when allowing refinement to run for 60 epochs. Most of the pipeline time is spent on embeddings optimization. \rrev{In the interactive mode, where the user marks additional constraints to improve the given clusters, we only re-run the Siamese network part of the pipeline, for a smaller amount of epochs. The refinement phase is tunable for a speed-accuracy tradeoff, and each optimization epoch takes about 0.5 seconds. For a common document of ~300 words, a refinement of 10 epochs, which we found was sufficient for convergence, would take up to 5 seconds. We include a detailed latency report in our supplementary material.}

\section{Results and Evaluation}
\label{sec:results}

\input{figures/user_study_misc/misc.tex}

We evaluated our technique on a new public dataset that we provide, of 50 highly varying images that have complicated layouts and span a wide range of documents, including menus, schedules, brochures and maps.
To quantitatively evaluate the quality of our propagation results, we conducted a user study. Additionally, we performed a comparison between our approach and previous works in document-image analysis, clarifying the gap between the objectives of these works and our objective. We also performed a qualitative ablation study, demonstrating the importance of the multimodal approach and the optimization scheme. Finally, we demonstrate generalization capabilities onto nearby domains.
In what follows, we elaborate on each of these experiments. We then conclude by discussing limitations of our presented approach. 

\subsection{User Study}
\label{sec:study}
We conducted a user study to quantitatively evaluate the quality and efficiency of our edit propagation system. As a semantic grouping is not entirely a well-defined task in the general case, we defined known categories and asked users to mark manual constraints and re-run the optimization until these known categories are grouped correctly. \rrev{We quantitatively compare the number of manual constraints defined by the user against manual approaches, counting the number of selections users would need to mark without our framework.}

We collected three sets of images capturing (i) menus, (ii) simple and (iii) dense document-images, where each set contains $10$ images in total. We defined two categories of interest in each set: items and item descriptions. In the case of menus, these are the items offered at the restaurant, and the corresponding item descriptions. Simple document-images are relatively sparse and for the most part contain these two categories only. Dense document-images capture mostly academic manuscripts or magazines, and in this case, the items are defined to be the paragraph titles and the item descriptions are defined to be the paragraphs of text. 
See Figure \ref{fig:user_misc} for a few samples presented in our study, along with our corresponding automatic grouping result. For the full image sets used in the study, please refer to the supplementary material.

The participants were first presented with instructions on how to select must-link and cannot-link constraints in our interactive system.  Thirty users in the age range 26-50 participated in the study (with a 1.14 male to female ratio). Each participant was provided with $10$ images belonging to one specific set and asked to evaluate the grouping of only one of the two categories. They were asked to add constraints and re-run the optimization until the grouping of that category was perfect. In other words, the users marked constraints until all the items/item descriptions were selected in a single group and no other words were included in that group. 

\input{figures/intro_fixed_classes/intro_fixed.tex}

We evaluated the quality of our automatically obtained result and of our interactive system by counting the number of user scribbles selected as constraints. By definition, a must-link constraint requires at least one scribble and specifies a group of words which should be clustered together, and a cannot-link constraint requires at least two scribbles and specifies two groups of words which should not be clustered together. 
To demonstrate that our system provides a more efficient way to propagate edit operations over simple counting mechanisms, we compared the number of manual constraints needed to three baselines that count the number of selections needed to group the words in the category. The most naive baseline counts the number of words in the category. The second baseline counts the number of horizontal lines in the category. Lastly, the third baseline counts the number of rectangular drag boxes needed to fully mark the category of interest; this baseline is proposed as editing softwares could optionally offer this functionality.     

\rrev{
Table \ref{tab:study} summarizes the results of our user study. For each set, we average over the results, where for each image we use the maximum number of words and scribbles needed to correctly group the predefined category. For example, on average, 0.5 user-scribbles are needed to fix the grouping of the menu item descriptions, while 8.6 rectangular drag boxes would be needed to achieve this grouping manually from scratch (which amount  to a selection of 118.9 words on average).

As the table illustrates, on average less than one scribble was needed to fix the initial grouping result. In contrast, more than five rectangular drag boxes are needed to mark the words in the predefined category, thus suggesting that our system can allow for a more efficient solution in selecting a category of words in a document-image. In the supplementary material, we provide all the images in the user-study along with the full comparison.  
}

\subsection{Comparison to previous work on document-images}
\label{sec:compare}

Previous works analyzing document-images assume a supervised setting and aim at segmenting the document image into a fixed set of labeled regions (e.g., section heading or paragraph). To illustrate the 
gap to our style- and content- aware edit propagation technique, in Figure \ref{fig:intro_fixed}, we compare our results to the segmentation results obtained by Yang et al. \shortcite{yang2017learning}.

As the figure illustrates, a segmentation technique which learns a predefined set of labels cannot allow for a fine-partitioning of the words within the image. See, for example, the names and job titles on the top row of Figure \ref{fig:intro_fixed}. They are both marked  as ``caption'', although semantically and visually these regions are not the same.

\input{figures/ablation/ablation_no_font_nlp.tex}

\input{figures/ablation/training/training.tex}

\subsection{Qualitative Ablation Results}
\label{sec:ablation}

To demonstrate the importance of the various stages in our pipeline, we provide qualitative ablation results and demonstrate that removing components from our solution decreases its quality substantially for various use cases. To further assess the results, please refer to the supplementary materials where we provide interactive ablation results for dozens of document-images, spanning a wide range of categories \rev{as well as results obtained in a supervised setting, training on multiple documents.}

\subsubsection*{Affinity space optimization}
We illustrate the importance of the optimization stage in our algorithm, which includes the training process of the Siamese Neural Network. To assess the quality of the grouping results without the optimization stage, we perform the following modifications to our pipeline: we use Kernel-PCA to project the high-dimensional representation to a lower dimensional space (setting the output dimensionality to $d_{out}$). To cluster the projected representations, we use an edge-weighted graph as described in Section \ref{sec:clustering}. However, as the neural net is unavailable, we employ a different predicate for the edge weights, using normalized Euclidean distance between pairs of words. For each pair of contextual lines, all pairwise distances of the enclosed words are averaged to yield the final affinity weight. 

In Figures \ref{fig:no_optimization}, \ref{fig:ablation_optimization1} and \ref{fig:overview} we visualize the non-learned affinity space. As the figures illustrate, unlike the learned affinity space, the non-learned affinity space is not tightly clustered. As the examples in the supplementary material illustrate, the clustering distance threshold for this case is unstable, and the quality is significantly lower compared to our constraint-based affinity learning approach. 

\subsubsection*{Multimodal representations}
Both the space of word representations in our work, as well as the constraints we employ in order to map this space into clusters, rely on a multitude of embedding types, namely: style, content and geometry. We recognize that for some document domains, specific embedding types may be more useful than others, and therefore demonstrate some use cases showing the effect of removing the aforementioned information from our representations and constraints. Not surprisingly, the visual style seems to be the most effective cue, and in some cases, a visual analysis is sufficient in correctly grouping the words in the document-image.  
In Figure \ref{fig:ablation_font_nlp}, we demonstrate results obtained using only visual style, style and content, style and geometry, and our full multimodal configuration.

In general, we argue that content embeddings enable our algorithm to differentiate between clusters sharing similar style traits, and yet have separate distinctive semantic properties. See, for example, the prices on the top of Figure \ref{fig:ablation_font_nlp}, which is grouped together with the item names when the content information is missing. We also show that geometric attributes (size and location of words) are useful to complement the visual description, in particular for the scenario where titles, headers and footers should be separated from other visually akin document elements harboring notably different functions. See, also, the map on the bottom of \ref{fig:ablation_font_nlp}, where the configuration without geometry yields a grouping where the seas and some of the countries and cities are grouped together. In this example, our full configuration best separates the seas, countries and cities into distinct clusters. 

\input{figures/scene_text/scene_new.tex}
\clearpage
\input{figures/scanned/scanned.tex}

\subsection{Generalization Capabilities}
\label{sec:gen}
\changes{
Next, we show results on more input types, without finetuning any parameters mentioned above, to demonstrate the generalization capabilities of our approach. 

We examine our method on images \emph{in the wild} that contain text (see Figure \ref{fig:scene_new}). While our propagation approach can more significantly boost performance in images that contain dozens of words, the editing operations can be performed on these images as well. \rev{It is important to note that, as our focus is on document-images, our solution is restricted for a 2D planar scenario. This implies, for instance, that our method will not cluster together text elements on two billboards at different depths in an image. 
}

We also demonstrate results on scanned documents, that contain various artifacts, including blur, distortions, etc. (see Figure \ref{fig:scanned}). These types of images are not only visually challenging, but also semantically, as the transcription is often erroneous. Nonetheless, our method is able to detect most of the unique clusters.

Furthermore, while our work focuses mostly with input images, containing rasterized documents, in the supplementary material we also demonstrate results of our algorithm on PDF documents, due to the proximity of this domain. The space of typical PDF documents can be roughly classified to documents with embedded images, which behave in a similar way to the document-images we process (text localization and transcription is not available in the input), as is usually the case with scanned documents.
The other common group is PDF documents containing text elements. The latter renders the OCR step of our algorithm unnecessary, as the text information can be extracted directly from the input. Although the representation of basic units may be further enriched with meta-data excavated from the digital document, meaningful clusters must still be inferred by sophisticated means, which in turn makes the remainder of our pipeline necessary.

}
\input{figures/limitations/limitations.tex}
\subsection{Limitations}
As a first step, our approach extracts the content and geometry of words using an off-the-shelf OCR solution. Thus, we are bounded by the accuracy of the OCR solution: errors from the OCR extraction phase are not fixed in later phases. OCR-related errors are evident in most figures, e.g., in Figure \ref{fig:scene_new}. \rev{While our algorithm is robust to these errors in most cases, in some cases, entire clusters are missed. For instance, as demonstrated in Figure \ref{fig:ablation_optimization}, our OCR engine fails at recognizing curved text. However, as available OCR solutions scale up to these more challenging scenarios, they will immediately be handled by the rest of our pipeline.  }

To augment the word affinities with a contextual notion, we defined contextual-lines in the image (Section \ref{sec:units}). Throughout our analysis, the contextual-lines serve as a basic building block, which we leverage multiple times. As we demonstrate, the contextual-lines allow for strengthening the initial multimodal representation. Furthermore, we use these lines to extract reliable pairwise connections from words within them in our optimization scheme, and ultimately propagate edits by grouping them.

Clearly, mistakes in the contextual-lines will lead to significant mistakes in the propagation result, which can only be corrected by user intervention. While our contextual-lines are defined conservatively and we deliberately avoid using any pretrained features in their selection, such mistakes are unavoidable. In Figure \ref{fig:limitations} we highlight an example, where the contextual-lines yield an undesirable grouping result as it merges two distinct regions.

Another notable limitation of our approach is that we analyze only the textual regions in the images, but in fact, a lot of documents contain regions of interest such as arrows, image sketches, or nested natural images. These may also assist in learning the affinities between the textual elements, and can even be considered as basic elements on their own right.

%% file: figures/architecture/architecture.tex
\begin{figure}

	\centering%
  \includegraphics[width=1.0\columnwidth]{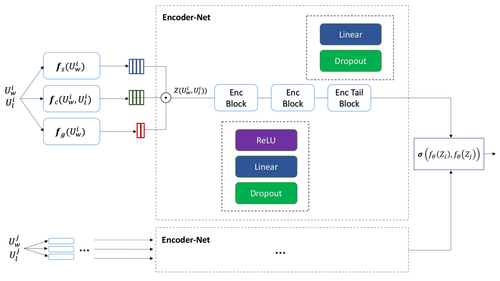}

\caption{ Siamese encoder network architecture. The input of the pipeline is a pair of words $U_w^i$, $U_w^j$ and their respective contextual-lines $U_l^i$, $U_l^j$. Multimodal representations are obtained for each word and are then concatenated to form the input to the network $Z(U_w, U_l)$. The network is composed of two Dropout-Linear-ReLU blocks, followed by a Dropout-Linear block. Finally, we compute the affinity of the two projected multimodal embeddings %
using the affinity function $\sigma$ as described in Section \ref{eq:affinity-final}. }
\label{fig:architecture}

\end{figure}

%% file: tables/user_study.tex
\begin{table}
\centering%
\vspace{-5pt}
\ra{1.0}
\setlength{\tabcolsep}{1.99pt}
\begin{tabular}{@{}llcccccccccccccccccc@{}}
\toprule
 \emph{Selection} && \multicolumn{2}{c} {\textbf{Menus}} && \multicolumn{2}{c} {\textbf{Simple}} && \multicolumn{2}{c} {\textbf{Dense}} && \multicolumn{2}{c} {\textbf{Average}} \\
  \emph{Method} && Items & Desc. && Items & Desc. && Items & Desc. && Items & Desc. \\ \midrule
\#words && 31.8 & 118.9 && 20.8 & 66.7 && 16.9 & 394.3 && 21.7 & 193.3 \\
  \#lines && 13.8 & 26.2 && 10.8 & 19.1 && 4.6 & 48.7 && 9.7 & 31.3 \\
  \#boxes && 8.3 & 8.6 && 4.7 & 4.7 && 4.1 & 5.1 && 5.7 & 6.1 \\
  ours && \textbf{2.0} & \textbf{0.5} && \textbf{0.5} & \textbf{0.3} && \textbf{0.4} & \textbf{0.3} && \textbf{1.0} & \textbf{0.3} \\
\bottomrule
\end{tabular}
\caption{\rrev{The user study results, comparing the number of post-optimization user edits to manual selection methods. 
We compare the number of user-scribble constraints needed using our approach against three baseline counting mechanisms, counting the number of words (\#words), lines (\#lines) and rectangular drag boxes (\#boxes) needed to group the corresponding categories items and item descriptions (Desc.) correctly. Lower is better. }
\label{tab:study}
}
\end{table}

%% file: figures/user_study_misc/misc.tex
\begin{figure*}
	\centering%
    \jsubfig{\fbox{\includegraphics[height=3.85cm]{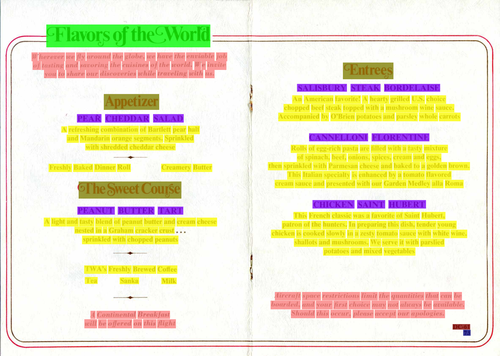}}}
	{}%
	\hfill    	\jsubfig{\fbox{\includegraphics[height=3.85cm]{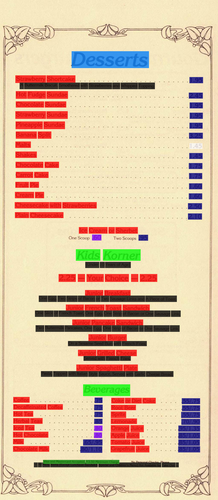}}}
	{}%
		\hfill
		\jsubfig{\fbox{\includegraphics[height=3.85cm]{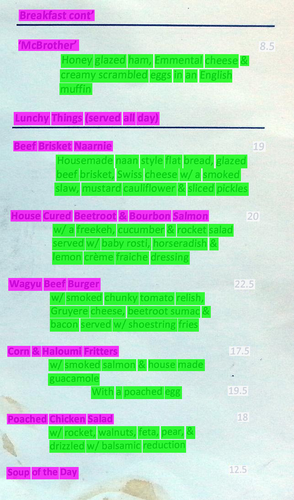}}}
	{}%
	\hfill    	\jsubfig{\fbox{\includegraphics[height=3.85cm]{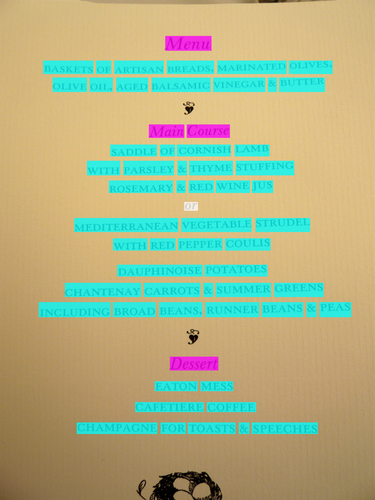}}}
	{}%
		\hfill
		\jsubfig{\fbox{\includegraphics[height=3.85cm]{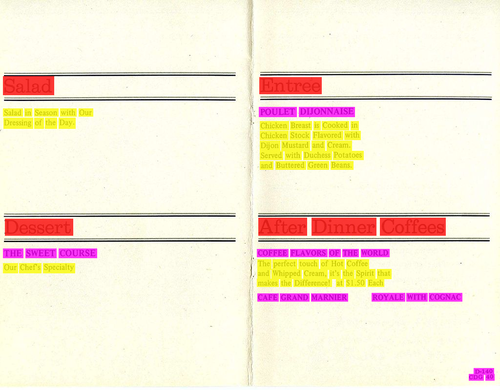}}}
	{}%
	\\
		\vspace{2pt}
    	\jsubfig{\fbox{\includegraphics[height=3.84cm]{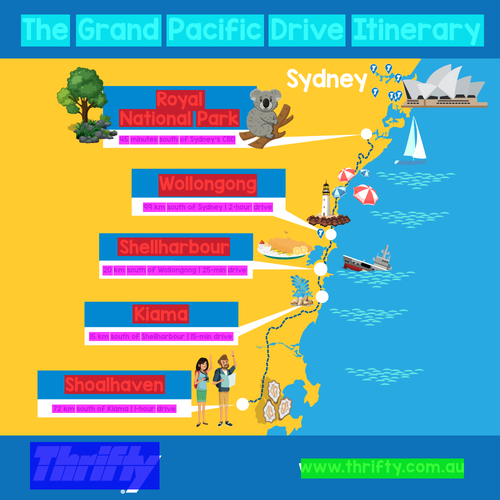}}}
	{}%
		\hfill
    	\jsubfig{\fbox{\includegraphics[height=3.84cm]{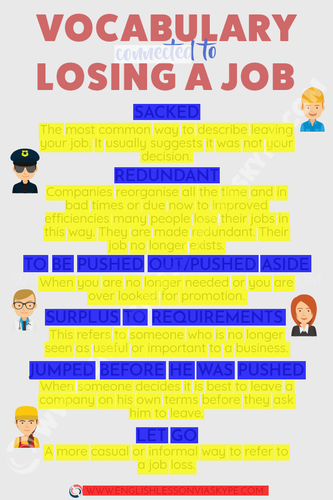}}}
    	{}
    		\hfill
    	\jsubfig{\fbox{\includegraphics[height=3.84cm]{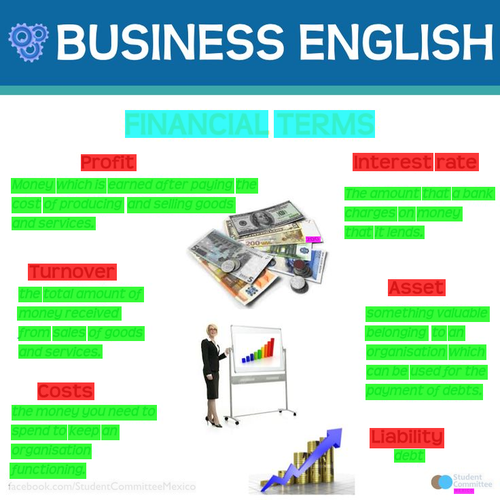}}}
    	{}
		\hfill
		\jsubfig{\fbox{\includegraphics[height=3.84cm]{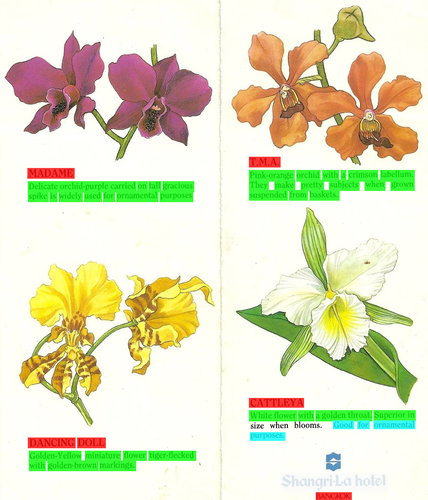}}}
    	{}
		\hfill
		\jsubfig{\fbox{\includegraphics[height=3.84cm]{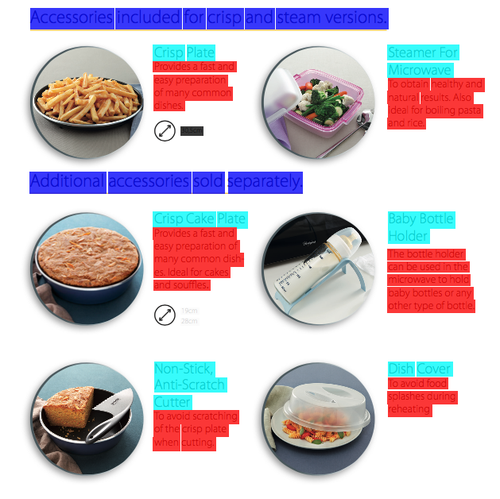}}}
    	{}
    	\\
    	\jsubfig{\fbox{\includegraphics[height=3.8cm]{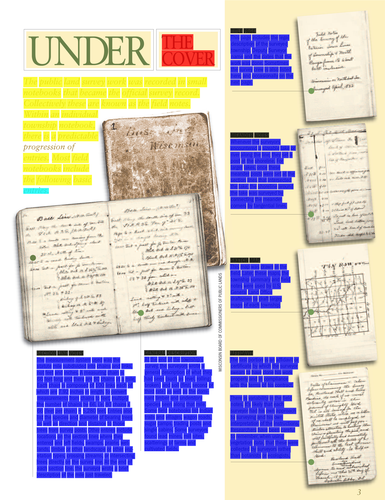}}}
    	{}
    			\hfill
    	\jsubfig{\fbox{\includegraphics[height=3.8cm]{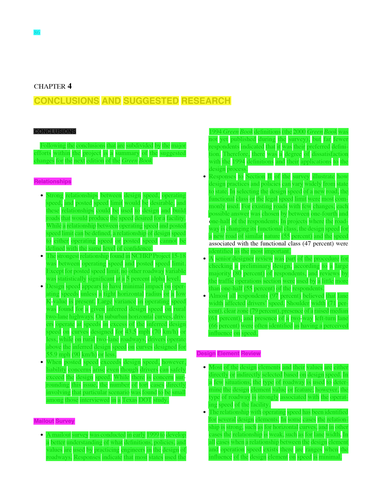}}}
    	{}
    	\hfill
    	\jsubfig{\fbox{\includegraphics[height=3.8cm]{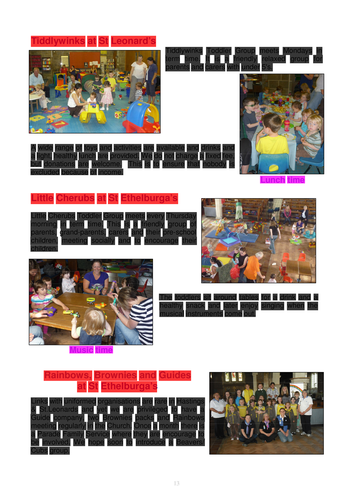}}}
    	{}
    	\hfill
    	\jsubfig{\fbox{\includegraphics[height=3.8cm]{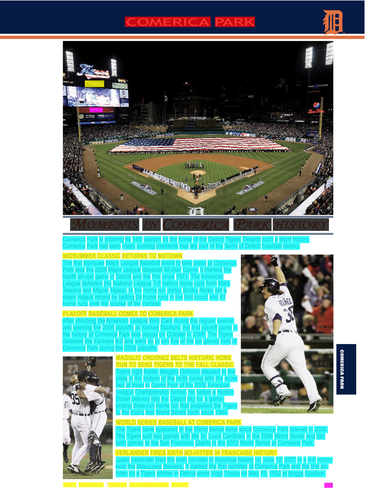}}}
    	{}
    	\hfill
    	\jsubfig{\fbox{\includegraphics[height=3.8cm]{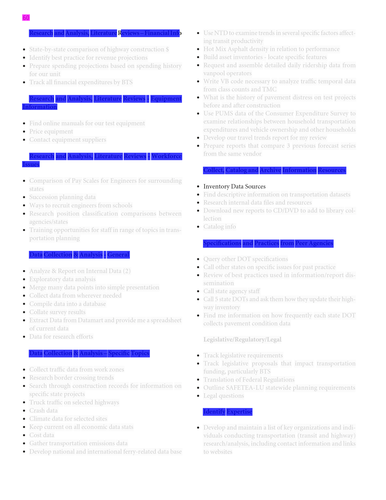}}}
    	{}
    	\hfill
    	\jsubfig{\fbox{\includegraphics[height=3.8cm]{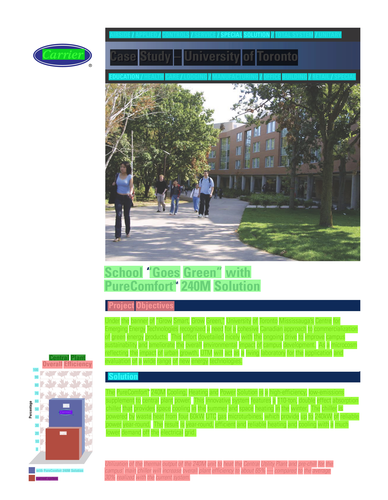}}}
    	{}

\caption{\textbf{User study image samples.} Users were presented with images capturing menus (top row), simple documents (middle row) or dense documents (bottom row). They were asked to manually mark constraints until the predefined category, items or item descriptions (paragraph titles and paragraphs in the case of dense documents), were grouped correctly. Above we illustrate a few samples with our automatically obtained grouping, which were later refined by the users. Please refer to the supplementary material for the full image sets used in the study.
}
\label{fig:user_misc}

\end{figure*}

%% file: figures/intro_fixed_classes/intro_fixed.tex
\begin{figure}
	\centering%
    	\jsubfig{{\includegraphics[height=7.5cm,trim={0 0 0 14.5cm},clip]{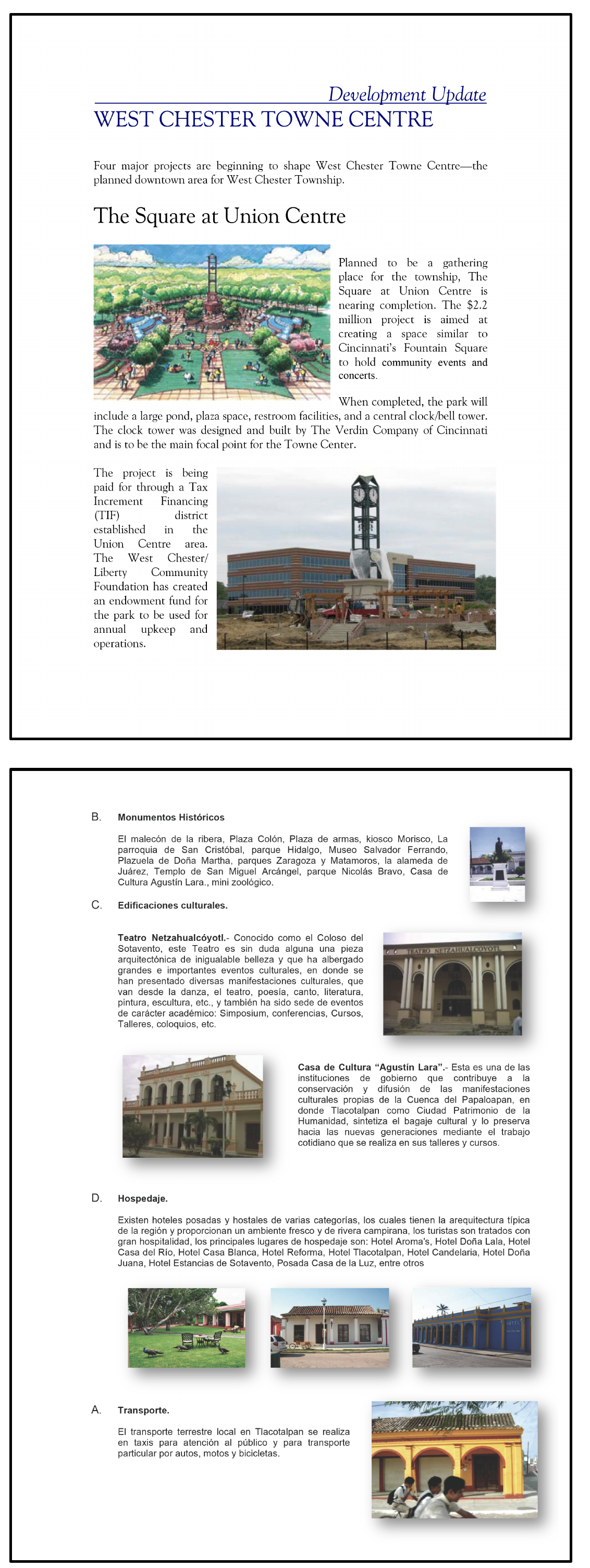}}}
	{Input}%
	\hfill%
    	\jsubfig{{\includegraphics[height=7.5cm,trim={0 0 0 14.5cm},clip]{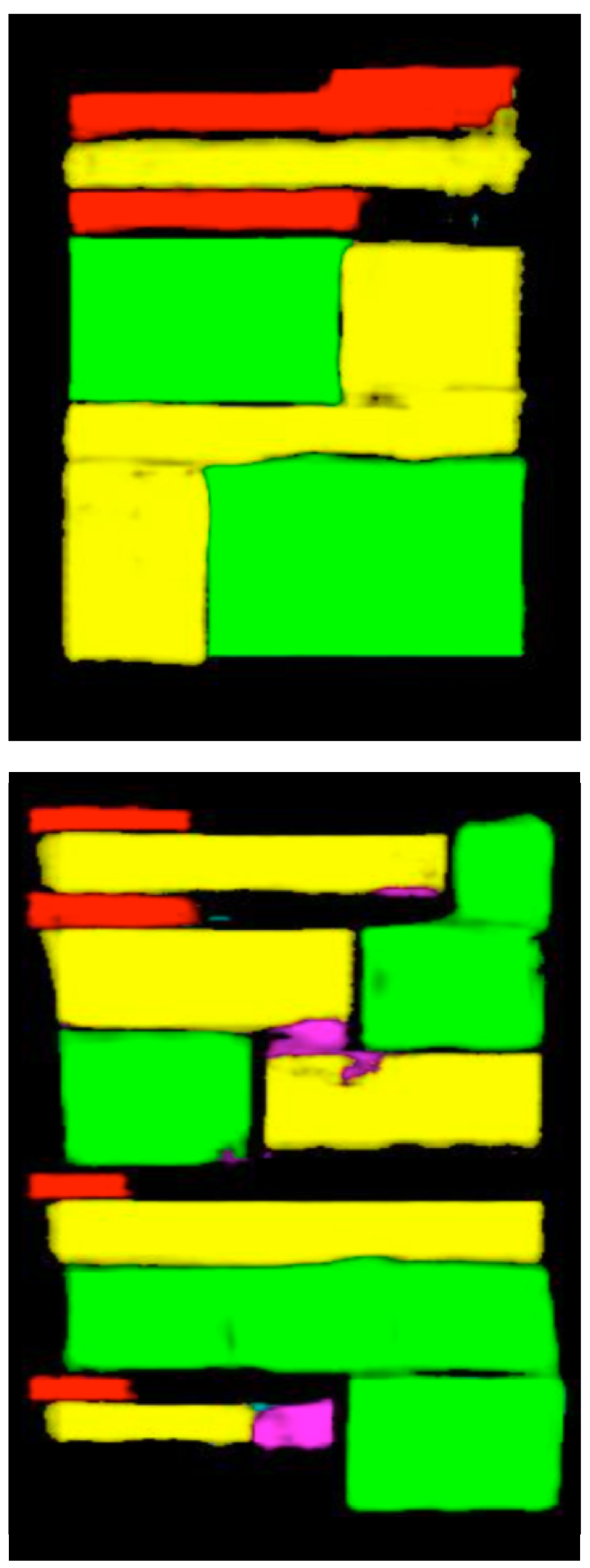}}}
	{\cite{yang2017learning}}%
	\hfill%
    	\jsubfig{{\includegraphics[height=7.5cm,trim={0 0 0 14.5cm},clip]{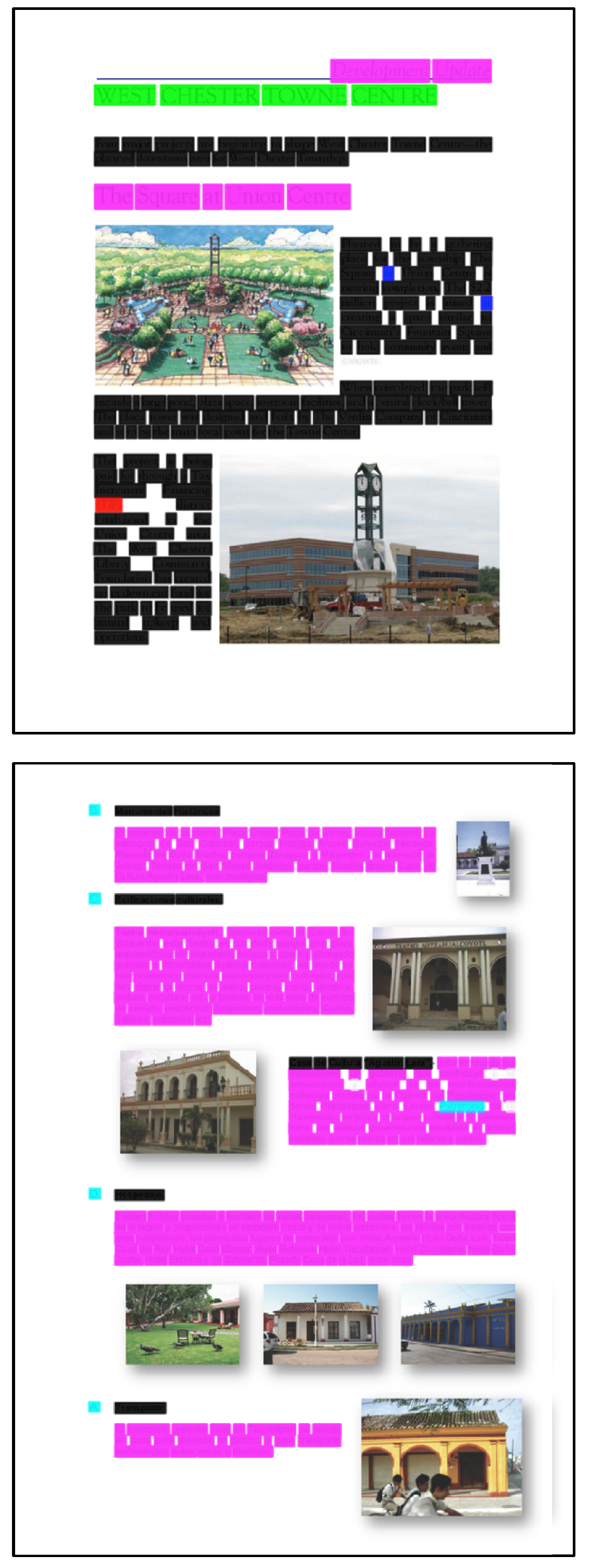}}}
	{Ours}%
	
\caption{\textbf{Comparison to previous work.} Previous work segment a document image into labelled regions (\rev{color legend:} \green{\textbf{figure}}, \todo{\textbf{section heading}}, \pink{\textbf{caption}}, and \yellow{\textbf{paragraph}}). While a general document structure can be inferred from these predefined classes, they do not allow for recovering the semantic affinities among words in the image. Semantically and visually unique regions may be merged into one labeled region (for example, the names and job titles in the top row). Our result is illustrated on the right, where the semantic groups are marked in unique colors.    }
\label{fig:intro_fixed}

\end{figure}

%% file: figures/ablation/ablation_no_font_nlp.tex
\begin{figure*}
	\centering%
		\jsubfig{\fbox{
	\includegraphics[height=2.65cm]{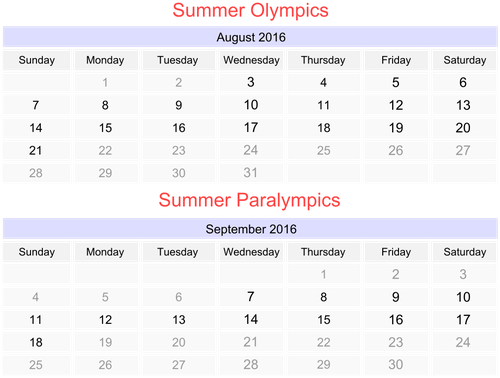}}}
	{}%
	\hfill
	\jsubfig{\fbox{
	\includegraphics[height=2.65cm]{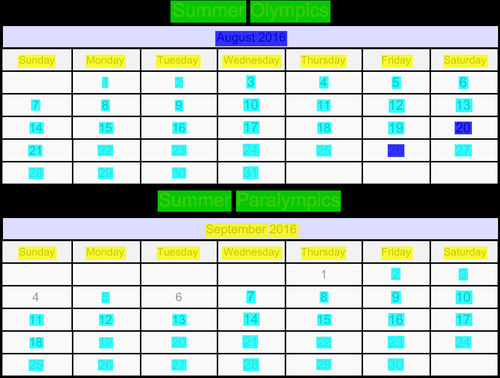}}}
	{}%
	\hfill
    \jsubfig{\fbox{\includegraphics[height=2.65cm]{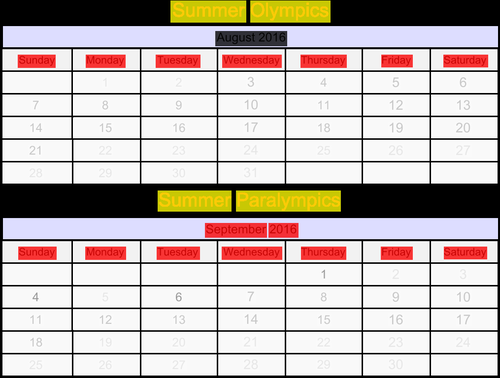}}}
	{}%
	\hfill
    \jsubfig{\fbox{\includegraphics[height=2.65cm]{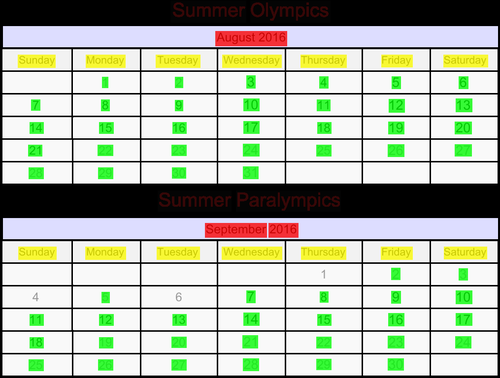}}}
	{}%
	\hfill
    \jsubfig{\fbox{\includegraphics[height=2.65cm]{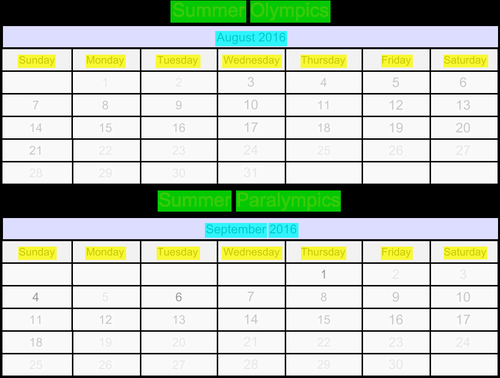}}}
	{ }%
	\\
\jsubfig{\fbox{
	\includegraphics[height=4.54cm]{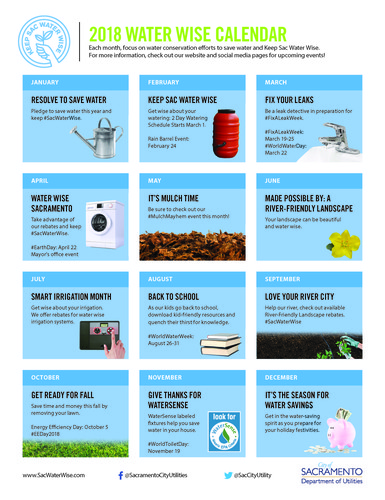}}}
	{}%
	\hfill
	\jsubfig{\fbox{
	\includegraphics[height=4.54cm]{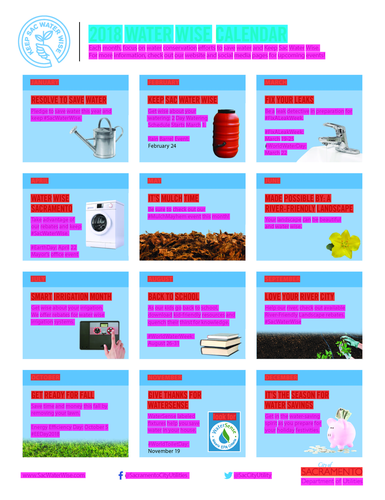}}}
	{}%
	\hfill
    \jsubfig{\fbox{\includegraphics[height=4.54cm]{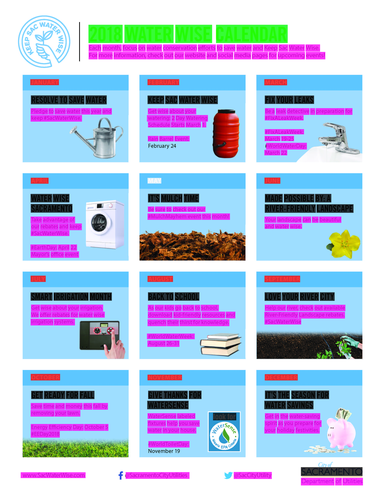}}}
	{}%
	\hfill
    \jsubfig{\fbox{\includegraphics[height=4.54cm]{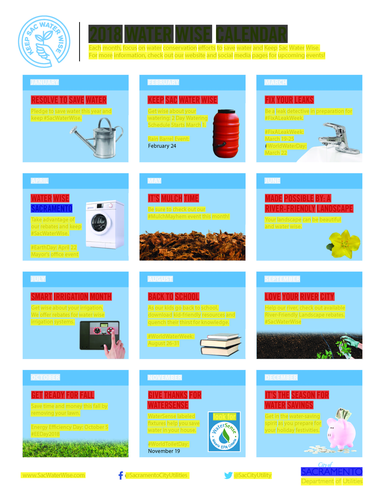}}}
	{}%
	\hfill
    \jsubfig{\fbox{\includegraphics[height=4.54cm]{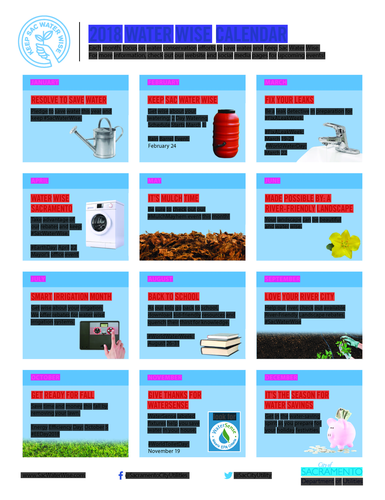}}}
	{  }
		\\
	\jsubfig{\fbox{
	\includegraphics[height=2.05cm]{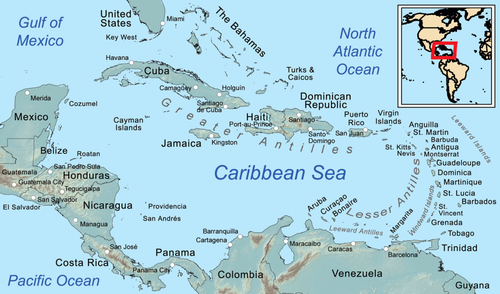}}}
	{Input}%
	\hfill
	\jsubfig{\fbox{
	\includegraphics[height=2.05cm]{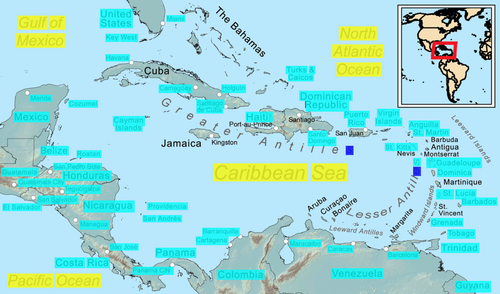}}}
	{Style \emph{Only}}%
	\hfill
    \jsubfig{\fbox{\includegraphics[height=2.05cm]{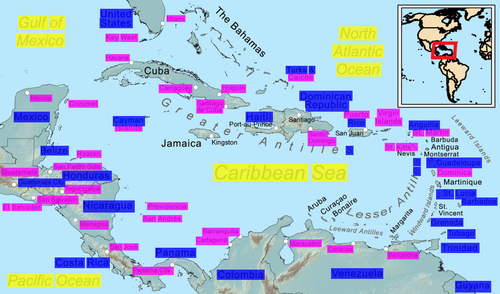}}}
	{\emph{Without} Content}%
	\hfill
    \jsubfig{\fbox{\includegraphics[height=2.05cm]{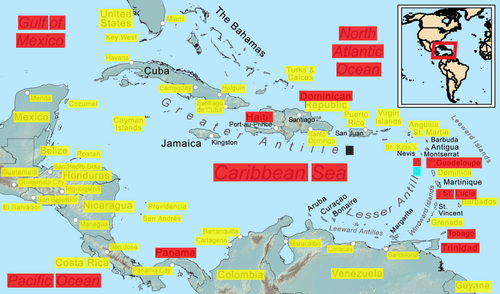}}}
	{\emph{Without} Geometry}%
	\hfill
    \jsubfig{\fbox{\includegraphics[height=2.05cm]{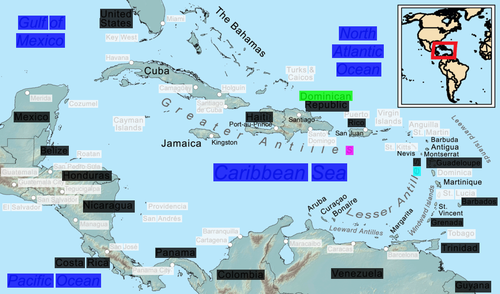}}}
	{\emph{Full} Algorithm  }%
\caption{\textbf{Ablation analysis of our multimodal approach.} We illustrate our grouping result with various multimodal configurations: only style (second to left), style and geometry (center), style and content (second to right), and our full multimodal configuration (left). Please refer to the supplementary materials for many more examples.  Original images due to (top to bottom): © Felipe Menegaz, © City of SACRAMENTO, © Karl Musser.
}
\label{fig:ablation_font_nlp}

\end{figure*}

%% file: figures/ablation/training/training.tex
\begin{figure}
\centering
			\jsubfig{\fbox{
	\includegraphics[height=3.98cm]{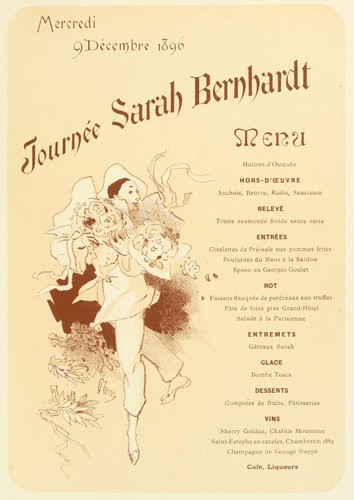}}}
	{Input}%
    \jsubfig{\fbox{\includegraphics[height=3.98cm]{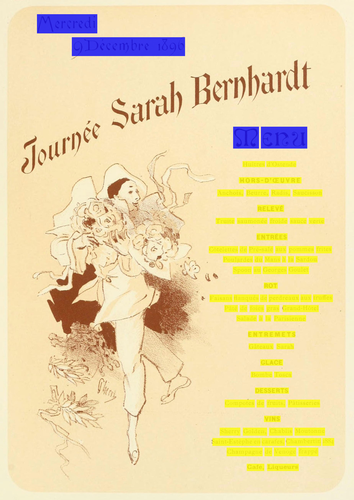}}}
	{\emph{Without} Learning}%
    \jsubfig{\fbox{\includegraphics[height=3.98cm]{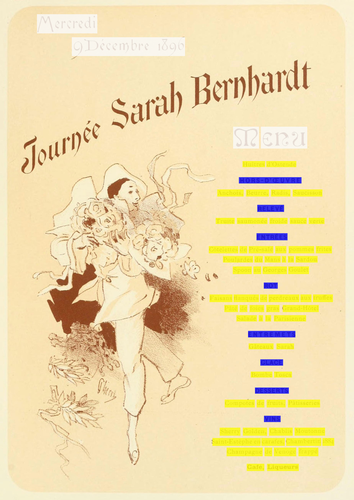}}
    }
	{\emph{Full} Algorithm }%
\caption{\textbf{Ablation analysis of our constraint-based affinity learning approach.} We illustrate our grouping result with and without our optimization scheme. %
As the figure illustrates, without our optimization scheme only some of the words are grouped correctly. Please refer to the supplementary materials for many more examples. Original image © Double--M, Menu de la journée Sarah Bernhardt, le 9 décembre 1896.
}
\label{fig:ablation_optimization}

\end{figure}

%% file: figures/scene_text/scene_new.tex
\begin{figure}
    \includegraphics[width=0.49\columnwidth]{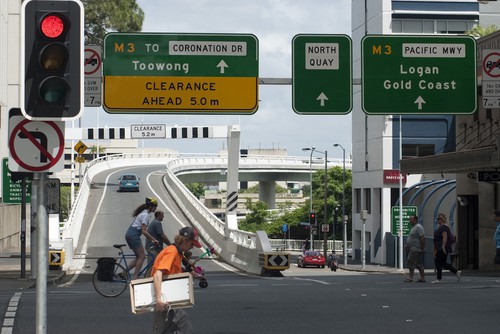}
     \includegraphics[width=0.49\columnwidth]{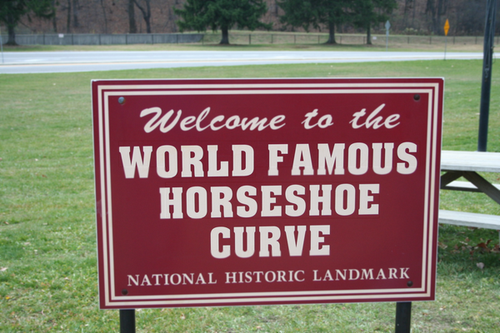}\\
         \includegraphics[width=0.49\columnwidth]{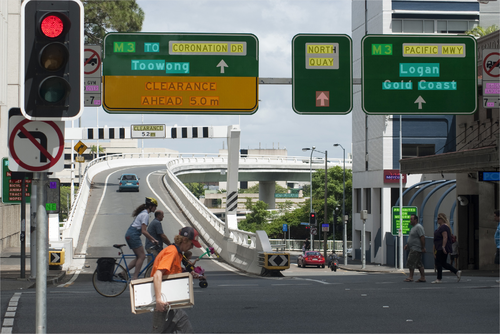}
     \includegraphics[width=0.49\columnwidth]{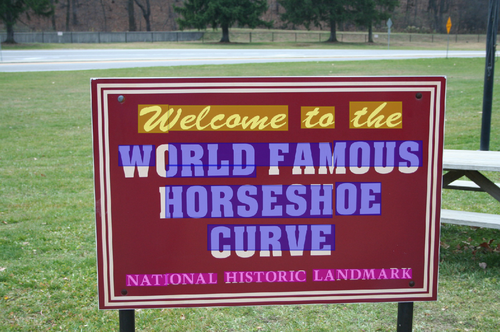}
	
\caption{Our presented technique is also applicable for images captured \emph{in the wild}, which contain only a small set of textual elements. Image courtesy Ron Shawley.
}
\label{fig:scene_new}

\end{figure}

%% file: figures/scanned/scanned.tex
\begin{figure}
    \centering
     \includegraphics[width=1.0\columnwidth]{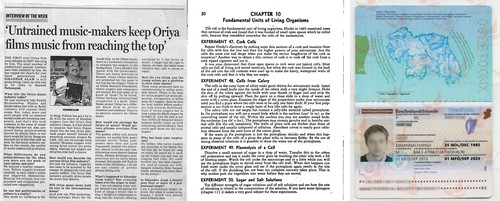} \\
     \includegraphics[width=1.0\columnwidth]{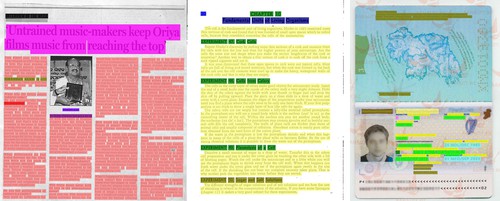}
    \caption{\changes{Learning affinities on scanned documents, which introduce various visual nuisances, as demonstrated above.}}
    \label{fig:scanned}
\end{figure}

%% file: figures/limitations/limitations.tex
\begin{figure}
	\centering%
		\jsubfig{
	\includegraphics[height=4.58cm]{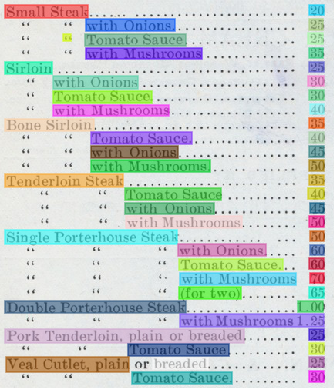}}
	{}%
	\hfill
    	\jsubfig{\includegraphics[height=4.58cm]{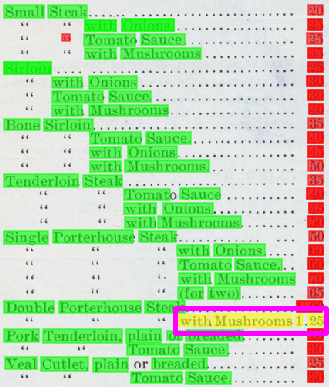}}
	{}%
\caption{\textbf{Contextual-line related limitations.} Our technique combines textual entities into the same contextual-line if the horizontal spacing between these entities is sufficiently small. However, as illustrated here, small horizontal space is not always a good indication of similar context.  
}
\label{fig:limitations}

\end{figure}

%% file: applications.tex
\input{figures/content_app/content_app.tex}
\input{figures/style_app_highlight/style_app.tex}

\section{Applications}
In this section, we demonstrate the usefulness of our approach, presenting several novel text editing applications based on our grouping technique. The edit operations are propagated to textual elements all across the image, thus significantly speeding-up the editing process. As previously discussed, editing textual content in a document-image can be a tedious task, especially if the amount of text entities is large, as the editing operations are repetitive in many cases. Below we illustrate examples of such potentially repetitive tasks. Our examples span style-based, content-based and geometry-based editing operations. Please refer to the accompanying video to view live editing examples.  

\paragraph{Style-based Applications} 
\label{sec:app_style}
Style-manipulation in images is a well-studied problem, especially recently due to advances in style transfer techniques (e.g., \cite{gatys2016image,zhu2017unpaired}). Within the context of local editing, some previous works focus on modifying the style of a region which captures a specific class, for example, a person \cite{shen2016automatic}.  

We are interested in modifying the style of all textual elements of a given type. We demonstrate two different style manipulation modes: emphasizing regions of interest and manipulating stylistic parameters, such as font type or color. 

In Figures \ref{fig:teaser} and \ref{fig:style_app_highlight}, we illustrate two filtering operations that can be used to emphasize selected groups within a document-image. We use standard image filters, including sharpness, contrast and brightness to create a highlighting effect on the selected group.  

To manipulate the stylistic parameters of the textual elements, we first need a mechanism to delete content within an image. For this purpose, we use a publicly available implementation of a standard image inpainting technique \cite{liu2018image} to delete content within a given bounding box region. To keep the font family parameter fixed, we use the full classifier model described in Section \ref{sec:app_style} to select the best matching font for editing.
In Figure \ref{fig:content_app}, we illustrate manipulation results of the font parameters (second image to the right). This tool can also be used to emphasize (or de-emphasize) regions of interest.

\paragraph{Content-based Applications}
Our technique can also facilitate propagating manipulations which are content-based. Indeed, our approach allows for fast propagation of semantic edits. In Figure \ref{fig:content_app}, we illustrate the benefit of a semantic tool of this sort. 
As demonstrated in the figure, using our tool, one can easily and quickly apply direct mathematical operations to elements that represent numbers or dates.  

\paragraph{Geometry-based Applications}
Modifying the relative location and ordering of similarly looking text entities is a common text editing procedure in images. Figure \ref{fig:geometry_app} illustrates a typical use-case where the ability to seamlessly modify the document structure is useful.
\input{figures/geometry_app/geometry_app.tex}

\paragraph{Interactive Document-images}
Document-images may contain excessive information, that may disturb the viewer's focus. To reduce the visual load, we can use the mechanism previously described to delete certain content in the image, and bring it back interactively as the user hovers the mouse over a nearby textual region. This allows for a dynamic image, where the user can digest the content in a gradual manner, unlike a still image where the content appears all at once. See the accompanying video for a demonstration of an interactive document-image which can be quickly generated using our edit propagation technique.

%% file: figures/content_app/content_app.tex
\begin{figure*}

	\centering%
    	\jsubfig{\fbox{\includegraphics[height=3.42cm]{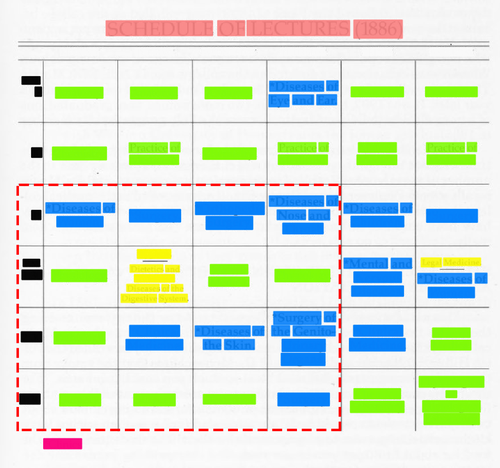}}}
	{}%
 	\hspace{1pt}
		\centering%
    	\jsubfig{\includegraphics[height=4.10cm]{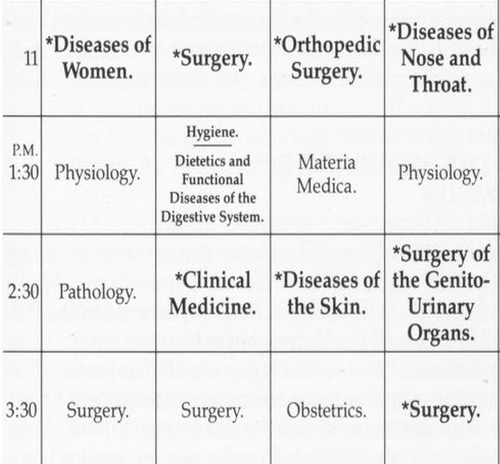}}
	{Original Image}%
		\centering%
		\hfill
    	\jsubfig{\includegraphics[height=4.10cm]{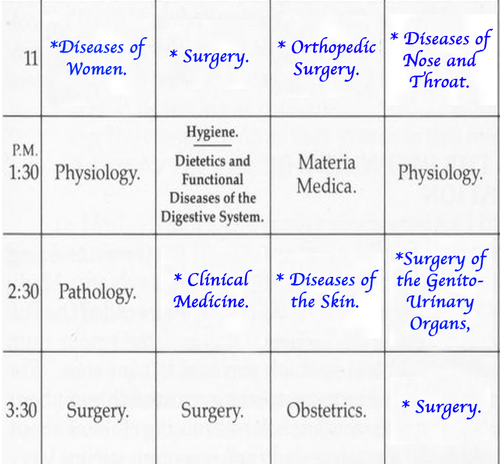}}
	{Style Modifications}%
		\centering%
 		\hspace{5pt}
    	\jsubfig{\includegraphics[height=4.10cm]{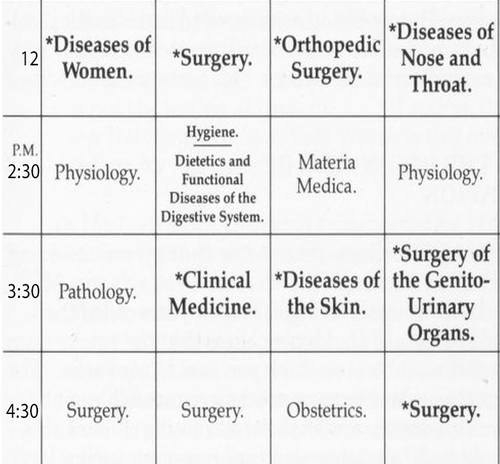}}
	{Content Modifications}%

\caption{\textbf{Style and Content Modifications.} We manipulate the style and the content of the our global textual elements (colored in unique colors on the left). Style manipulations include modifications to font characteristics and font color (second image to the right), and content manipulations include mathematical operations. On the rightmost image, we modify the schedule by adding $+1$ to all time entities; note the left column of this image.  
}
\label{fig:content_app}

\end{figure*}

%% file: figures/style_app_highlight/style_app.tex
\begin{figure}
	\includegraphics[width=1.0\columnwidth]{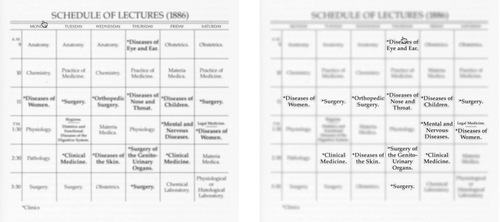}

\caption{\textbf{Emphasizing textual elements in the image.} Our technique enables propagating edits that emphasize textual elements of a certain type. For the extracted input image and the grouping result illustrated in Figure \ref{fig:content_app}, the user can click a single word to emphasize a particular category of textual elements within the image. 
}
\label{fig:style_app_highlight}

\end{figure}

%% file: figures/geometry_app/geometry_app.tex
\begin{figure}

	\centering%
    	\jsubfig{\includegraphics[height=3.02cm]{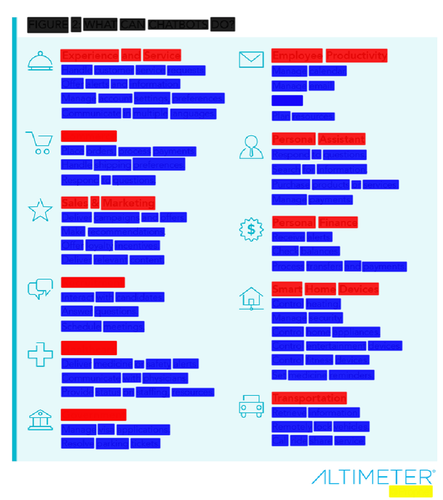}}
	{}%
	\hfill
		\centering%
    	\jsubfig{\includegraphics[height=3.02cm]{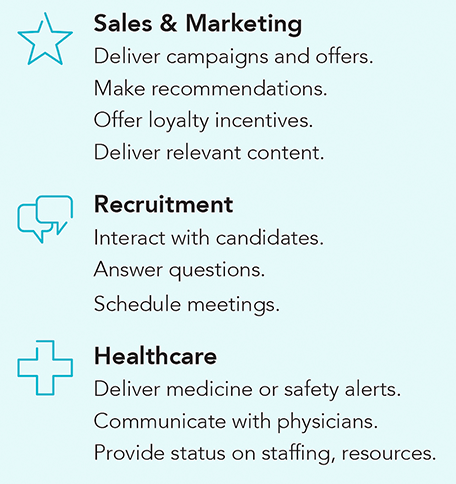}}
	{}%
		\centering%
		\hfill
    	\jsubfig{\includegraphics[height=3.02cm]{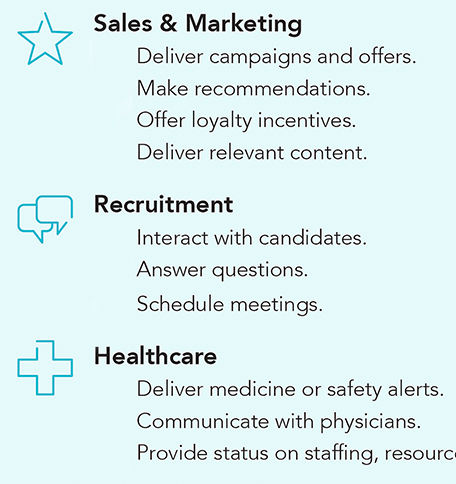}}
	{}%

\caption{\textbf{Structure Modifications.} Our technique enables manipulating the structure of global elements. Above, we modify the horizontal location of the group that is colored in blue (on the leftmost figure) by indenting the text within this textual element. Image courtesy: Altimeter @ Prophet.
}
\label{fig:geometry_app}

\end{figure}

%% file: conclusions.tex
\section{Conclusions and Future Work}

We have presented an unsupervised method to learn affinities between words in a document-image, where the affinity reflects not only closeness in visual appearance, but also in the semantics of the underlying text. We optimized a multimodal affinity space by leveraging reliable pairwise connections in the data. Our pairs-based approach allows for the integration of additional supervised pairs which further enhances the affinity space. Technically, to learn an appropriate affinity using the pairwise connections, we represented the textual elements in the image in a multimodal manner that captures the style, content, and geometry of each word; and then optimized our affinity space
with the help of a Siamese network, which embeds our multimodal word representation to a latent low-dimensional affinity space.

In a sense, our technique of grouping via unsupervised affinity learning can be seen as a way to \emph{categorize} the textual elements in the image into several groups, or categories, that are learned in an unsupervised manner. This diverges significantly from most of the previous works on document analysis that usually partition the documents into categories labeled in advance, such as \emph{title}, \emph{paragraph title}, \emph{paragraph}, and \emph{caption}. The fact that our categories are learned in an unsupervised manner is essential to deal with various types of documents without prior knowledge on the document type and characteristics, and indeed, we show that our method performs well on many document types, including dense text documents, menus, calendars, maps, and brochures. The unsupervised grouping is shown to be useful in accelerating edit propagation, but in fact, it might serve as a new, generic way to represent and perceive documents of any type in a way that is similar to the human perception, and useful for document understanding applications.

In the future, we would like to further explore the relations and hierarchy between grouped textual elements in the image, tackling the problem of reconstructing text layout and re-flowing text from an input image. Another interesting avenue for future research is a multimodal image retrieval methodology, used to obtain document-images with similar style and content.

One application where our approach might be particularly useful is in digitization of documents with complex layouts. Digitization is the process of converting ``real-world'' information into a reliable computer-readable format. Substantial efforts have been dedicated to digitization of dense documents like books \cite{COYLE2006} and newspapers or journals \cite{ICDAR2017}. However, we are not aware of any generic digitization tool that can extract the textual content from \emph{any} document while preserving the information encapsulated in the structure of the document. Such a tool would need to understand high level relations between entities that are not necessarily co-located, and our embedding and grouping techniques might readily serve as a useful building block for tasks of this type.

%% file: main.bbl

\begin{thebibliography}{00}

%%% the Web address.

\ifx \showCODEN    \undefined \def \showCODEN     #1{\unskip}     \fi
\ifx \showDOI      \undefined \def \showDOI       #1{#1}\fi
\ifx \showISBNx    \undefined \def \showISBNx     #1{\unskip}     \fi
\ifx \showISBNxiii \undefined \def \showISBNxiii  #1{\unskip}     \fi
\ifx \showISSN     \undefined \def \showISSN      #1{\unskip}     \fi
\ifx \showLCCN     \undefined \def \showLCCN      #1{\unskip}     \fi
\ifx \shownote     \undefined \def \shownote      #1{#1}          \fi
\ifx \showarticletitle \undefined \def \showarticletitle #1{#1}   \fi
\ifx \showURL      \undefined \def \showURL       {\relax}        \fi
% invisible to TeX
\providecommand\bibfield[2]{#2}
\providecommand\bibinfo[2]{#2}
\providecommand\natexlab[1]{#1}
\providecommand\showeprint[2][]{arXiv:#2}

\bibitem[\protect\citeauthoryear{Agrawal, Batra, Parikh, and Kembhavi}{Agrawal
  et~al\mbox{.}}{2018}]%
        {vqa-cp}
\bibfield{author}{\bibinfo{person}{Aishwarya Agrawal}, \bibinfo{person}{Dhruv
  Batra}, \bibinfo{person}{Devi Parikh}, {and} \bibinfo{person}{Aniruddha
  Kembhavi}.} \bibinfo{year}{2018}\natexlab{}.
\newblock \showarticletitle{Don't Just Assume; Look and Answer: Overcoming
  Priors for Visual Question Answering}. In \bibinfo{booktitle}{{\em IEEE
  Conference on Computer Vision and Pattern Recognition (CVPR)}}.
\newblock


\bibitem[\protect\citeauthoryear{Akbik, Blythe, and Vollgraf}{Akbik
  et~al\mbox{.}}{2018}]%
        {akbik2018coling}
\bibfield{author}{\bibinfo{person}{Alan Akbik}, \bibinfo{person}{Duncan
  Blythe}, {and} \bibinfo{person}{Roland Vollgraf}.}
  \bibinfo{year}{2018}\natexlab{}.
\newblock \showarticletitle{Contextual String Embeddings for Sequence
  Labeling}. In \bibinfo{booktitle}{{\em {COLING} 2018, 27th International
  Conference on Computational Linguistics}}. \bibinfo{pages}{1638--1649}.
\newblock


\bibitem[\protect\citeauthoryear{Akiyama and Hagita}{Akiyama and
  Hagita}{1990}]%
        {Akiyama1990AutomatedES}
\bibfield{author}{\bibinfo{person}{Teruo Akiyama} {and}
  \bibinfo{person}{Norihiro Hagita}.} \bibinfo{year}{1990}\natexlab{}.
\newblock \showarticletitle{Automated entry system for printed documents}.
\newblock \bibinfo{journal}{{\em Pattern Recognition\/}}  \bibinfo{volume}{23}
  (\bibinfo{year}{1990}), \bibinfo{pages}{1141--1154}.
\newblock


\bibitem[\protect\citeauthoryear{An and Pellacini}{An and Pellacini}{2008}]%
        {an2008appprop}
\bibfield{author}{\bibinfo{person}{Xiaobo An} {and} \bibinfo{person}{Fabio
  Pellacini}.} \bibinfo{year}{2008}\natexlab{}.
\newblock \showarticletitle{AppProp: all-pairs appearance-space edit
  propagation}. In \bibinfo{booktitle}{{\em ACM Transactions on Graphics
  (TOG)}}, Vol.~\bibinfo{volume}{27}. ACM, \bibinfo{pages}{40}.
\newblock


\bibitem[\protect\citeauthoryear{Antol, Agrawal, Lu, Mitchell, Batra, Zitnick,
  and Parikh}{Antol et~al\mbox{.}}{2015}]%
        {VQA}
\bibfield{author}{\bibinfo{person}{Stanislaw Antol}, \bibinfo{person}{Aishwarya
  Agrawal}, \bibinfo{person}{Jiasen Lu}, \bibinfo{person}{Margaret Mitchell},
  \bibinfo{person}{Dhruv Batra}, \bibinfo{person}{C.~Lawrence Zitnick}, {and}
  \bibinfo{person}{Devi Parikh}.} \bibinfo{year}{2015}\natexlab{}.
\newblock \showarticletitle{{VQA}: {V}isual {Q}uestion {A}nswering}. In
  \bibinfo{booktitle}{{\em International Conference on Computer Vision
  (ICCV)}}.
\newblock


\bibitem[\protect\citeauthoryear{Baird, Bunke, and Yamamoto}{Baird
  et~al\mbox{.}}{1992}]%
        {Baird1992StructuredDI}
\bibfield{author}{\bibinfo{person}{Henry~S. Baird}, \bibinfo{person}{Horst
  Bunke}, {and} \bibinfo{person}{Kazuhiko Yamamoto}.}
  \bibinfo{year}{1992}\natexlab{}.
\newblock \showarticletitle{Structured Document Image Analysis}. In
  \bibinfo{booktitle}{{\em Springer Berlin Heidelberg}}.
\newblock


\bibitem[\protect\citeauthoryear{Barnes, Shechtman, Finkelstein, and
  Goldman}{Barnes et~al\mbox{.}}{2009}]%
        {barnes2009patchmatch}
\bibfield{author}{\bibinfo{person}{Connelly Barnes}, \bibinfo{person}{Eli
  Shechtman}, \bibinfo{person}{Adam Finkelstein}, {and} \bibinfo{person}{Dan~B
  Goldman}.} \bibinfo{year}{2009}\natexlab{}.
\newblock \showarticletitle{PatchMatch: A randomized correspondence algorithm
  for structural image editing}.
\newblock \bibinfo{journal}{{\em ACM Transactions on Graphics (ToG)\/}}
  \bibinfo{volume}{28}, \bibinfo{number}{3} (\bibinfo{year}{2009}),
  \bibinfo{pages}{24}.
\newblock


\bibitem[\protect\citeauthoryear{Breuel}{Breuel}{2002}]%
        {Breuel2002}
\bibfield{author}{\bibinfo{person}{Thomas~M. Breuel}.}
  \bibinfo{year}{2002}\natexlab{}.
\newblock \showarticletitle{Two Geometric Algorithms for Layout Analysis}. In
  \bibinfo{booktitle}{{\em Document Analysis Systems V}},
  \bibfield{editor}{\bibinfo{person}{Daniel Lopresti},
  \bibinfo{person}{Jianying Hu}, {and} \bibinfo{person}{Ramanujan Kashi}}
  (Eds.). \bibinfo{publisher}{Springer Berlin Heidelberg},
  \bibinfo{pages}{188--199}.
\newblock


\bibitem[\protect\citeauthoryear{Breuel}{Breuel}{2003}]%
        {Breuel2003HighPD}
\bibfield{author}{\bibinfo{person}{Thomas~M. Breuel}.}
  \bibinfo{year}{2003}\natexlab{}.
\newblock \showarticletitle{High Performance Document Layout Analysis}. In
  \bibinfo{booktitle}{{\em Proceedings of the Symposium on Document Image
  Understanding Technology}}.
\newblock


\bibitem[\protect\citeauthoryear{Bylinskii, Alsheikh, Madan, Recasens, Zhong,
  Pfister, Durand, and Oliva}{Bylinskii et~al\mbox{.}}{2017}]%
        {visually1}
\bibfield{author}{\bibinfo{person}{Zoya Bylinskii}, \bibinfo{person}{Sami
  Alsheikh}, \bibinfo{person}{Spandan Madan}, \bibinfo{person}{Adria Recasens},
  \bibinfo{person}{Kimberli Zhong}, \bibinfo{person}{Hanspeter Pfister},
  \bibinfo{person}{Fredo Durand}, {and} \bibinfo{person}{Aude Oliva}.}
  \bibinfo{year}{2017}\natexlab{}.
\newblock \showarticletitle{Understanding infographics through textual and
  visual tag prediction}. In \bibinfo{booktitle}{{\em arXiv preprint
  arXiv:1709.09215}}.
\newblock
\showURL{%
\url{https://arxiv.org/pdf/1709.09215}}


\bibitem[\protect\citeauthoryear{Chen, Zou, Zhao, and Tan}{Chen
  et~al\mbox{.}}{2012}]%
        {chen2012manifold}
\bibfield{author}{\bibinfo{person}{Xiaowu Chen}, \bibinfo{person}{Dongqing
  Zou}, \bibinfo{person}{Qinping Zhao}, {and} \bibinfo{person}{Ping Tan}.}
  \bibinfo{year}{2012}\natexlab{}.
\newblock \showarticletitle{Manifold preserving edit propagation}.
\newblock \bibinfo{journal}{{\em ACM Transactions on Graphics (TOG)\/}}
  \bibinfo{volume}{31}, \bibinfo{number}{6} (\bibinfo{year}{2012}),
  \bibinfo{pages}{132}.
\newblock


\bibitem[\protect\citeauthoryear{Chen, Wang, Wang, Wang, and Qu}{Chen
  et~al\mbox{.}}{2020}]%
        {ChenWWWQ20}
\bibfield{author}{\bibinfo{person}{Zhutian Chen}, \bibinfo{person}{Yun Wang},
  \bibinfo{person}{Qianwen Wang}, \bibinfo{person}{Yong Wang}, {and}
  \bibinfo{person}{Huamin Qu}.} \bibinfo{year}{2020}\natexlab{}.
\newblock \showarticletitle{Towards Automated Infographic Design: Deep
  Learning-based Auto-Extraction of Extensible Timeline}.
\newblock \bibinfo{journal}{{\em {IEEE} Trans. Vis. Comput. Graph.\/}}
  \bibinfo{volume}{26}, \bibinfo{number}{1} (\bibinfo{year}{2020}),
  \bibinfo{pages}{917--926}.
\newblock


\bibitem[\protect\citeauthoryear{Clausner, Antonacopoulos, and
  Pletschacher}{Clausner et~al\mbox{.}}{2017}]%
        {ICDAR2017}
\bibfield{author}{\bibinfo{person}{C. Clausner}, \bibinfo{person}{A.
  Antonacopoulos}, {and} \bibinfo{person}{S. Pletschacher}.}
  \bibinfo{year}{2017}\natexlab{}.
\newblock \showarticletitle{ICDAR2017 Competition on Recognition of Documents
  with Complex Layouts - RDCL2017}, Vol.~\bibinfo{volume}{01}.
  \bibinfo{pages}{1404--1410}.
\newblock


\bibitem[\protect\citeauthoryear{Coyle}{Coyle}{2006}]%
        {COYLE2006}
\bibfield{author}{\bibinfo{person}{Karen Coyle}.}
  \bibinfo{year}{2006}\natexlab{}.
\newblock \showarticletitle{Mass Digitization of Books}.
\newblock \bibinfo{journal}{{\em The Journal of Academic Librarianship\/}}
  \bibinfo{volume}{32}, \bibinfo{number}{6} (\bibinfo{year}{2006}),
  \bibinfo{pages}{641 -- 645}.
\newblock


\bibitem[\protect\citeauthoryear{Dizaji, Herandi, Deng, Cai, and Huang}{Dizaji
  et~al\mbox{.}}{2017}]%
        {dizaji2017deep}
\bibfield{author}{\bibinfo{person}{Kamran~Ghasedi Dizaji},
  \bibinfo{person}{Amirhossein Herandi}, \bibinfo{person}{Cheng Deng},
  \bibinfo{person}{Weidong Cai}, {and} \bibinfo{person}{Heng Huang}.}
  \bibinfo{year}{2017}\natexlab{}.
\newblock \showarticletitle{Deep clustering via joint convolutional autoencoder
  embedding and relative entropy minimization}. In \bibinfo{booktitle}{{\em
  2017 IEEE International Conference on Computer Vision (ICCV)}}. IEEE,
  \bibinfo{pages}{5747--5756}.
\newblock


\bibitem[\protect\citeauthoryear{Donahue, Hendricks, Rohrbach, Venugopalan,
  Guadarrama, Saenko, and Darrell}{Donahue et~al\mbox{.}}{2017}]%
        {Donahue17}
\bibfield{author}{\bibinfo{person}{J. Donahue}, \bibinfo{person}{L.~A.
  Hendricks}, \bibinfo{person}{M. Rohrbach}, \bibinfo{person}{S. Venugopalan},
  \bibinfo{person}{S. Guadarrama}, \bibinfo{person}{K. Saenko}, {and}
  \bibinfo{person}{T. Darrell}.} \bibinfo{year}{2017}\natexlab{}.
\newblock \showarticletitle{Long-Term Recurrent Convolutional Networks for
  Visual Recognition and Description}.
\newblock \bibinfo{journal}{{\em IEEE Transactions on Pattern Analysis and
  Machine Intelligence\/}} \bibinfo{volume}{39}, \bibinfo{number}{4}
  (\bibinfo{year}{2017}), \bibinfo{pages}{677--691}.
\newblock


\bibitem[\protect\citeauthoryear{Farbman, Fattal, and Lischinski}{Farbman
  et~al\mbox{.}}{2010}]%
        {farbman2010diffusion}
\bibfield{author}{\bibinfo{person}{Zeev Farbman}, \bibinfo{person}{Raanan
  Fattal}, {and} \bibinfo{person}{Dani Lischinski}.}
  \bibinfo{year}{2010}\natexlab{}.
\newblock \showarticletitle{Diffusion maps for edge-aware image editing}. In
  \bibinfo{booktitle}{{\em ACM Transactions on Graphics (TOG)}},
  Vol.~\bibinfo{volume}{29}. ACM, \bibinfo{pages}{145}.
\newblock


\bibitem[\protect\citeauthoryear{Fathi, Wojna, Rathod, Wang, Song, Guadarrama,
  and Murphy}{Fathi et~al\mbox{.}}{2017}]%
        {Fathi2017SemanticIS}
\bibfield{author}{\bibinfo{person}{Alireza Fathi}, \bibinfo{person}{Zbigniew
  Wojna}, \bibinfo{person}{Vivek Rathod}, \bibinfo{person}{Peng Wang},
  \bibinfo{person}{Hyun~Oh Song}, \bibinfo{person}{Sergio Guadarrama}, {and}
  \bibinfo{person}{Kevin~P. Murphy}.} \bibinfo{year}{2017}\natexlab{}.
\newblock \showarticletitle{Semantic Instance Segmentation via Deep Metric
  Learning}.
\newblock \bibinfo{journal}{{\em CoRR\/}}  \bibinfo{volume}{abs/1703.10277}
  (\bibinfo{year}{2017}).
\newblock


\bibitem[\protect\citeauthoryear{Fogel, Averbuch-Elor, Goldberger, and
  Cohen-Or}{Fogel et~al\mbox{.}}{2018}]%
        {fogel2018clustering}
\bibfield{author}{\bibinfo{person}{Sharon Fogel}, \bibinfo{person}{Hadar
  Averbuch-Elor}, \bibinfo{person}{Jacov Goldberger}, {and}
  \bibinfo{person}{Daniel Cohen-Or}.} \bibinfo{year}{2018}\natexlab{}.
\newblock \showarticletitle{Clustering-driven Deep Embedding with Pairwise
  Constraints}.
\newblock \bibinfo{journal}{{\em IEEE Computer Graphics and Applications\/}}
  (\bibinfo{year}{2018}).
\newblock


\bibitem[\protect\citeauthoryear{Garces, Agarwala, Gutierrez, and
  Hertzmann}{Garces et~al\mbox{.}}{2014}]%
        {garces2014}
\bibfield{author}{\bibinfo{person}{Elena Garces}, \bibinfo{person}{Aseem
  Agarwala}, \bibinfo{person}{Diego Gutierrez}, {and} \bibinfo{person}{Aaron
  Hertzmann}.} \bibinfo{year}{2014}\natexlab{}.
\newblock \showarticletitle{A Similarity Measure for Illustration Style}.
\newblock \bibinfo{journal}{{\em ACM Transactions on Graphics (TOG)\/}}
  \bibinfo{volume}{33}, \bibinfo{number}{4} (\bibinfo{year}{2014}),
  \bibinfo{pages}{93:1--93:9}.
\newblock


\bibitem[\protect\citeauthoryear{Gatys, Ecker, and Bethge}{Gatys
  et~al\mbox{.}}{2016}]%
        {gatys2016image}
\bibfield{author}{\bibinfo{person}{Leon~A Gatys}, \bibinfo{person}{Alexander~S
  Ecker}, {and} \bibinfo{person}{Matthias Bethge}.}
  \bibinfo{year}{2016}\natexlab{}.
\newblock \showarticletitle{Image style transfer using convolutional neural
  networks}. In \bibinfo{booktitle}{{\em Proceedings of the IEEE Conference on
  Computer Vision and Pattern Recognition}}. \bibinfo{pages}{2414--2423}.
\newblock


\bibitem[\protect\citeauthoryear{Goodfellow, Pouget-Abadie, Mirza, Xu,
  Warde-Farley, Ozair, Courville, and Bengio}{Goodfellow et~al\mbox{.}}{2014}]%
        {goodfellow2014generative}
\bibfield{author}{\bibinfo{person}{Ian Goodfellow}, \bibinfo{person}{Jean
  Pouget-Abadie}, \bibinfo{person}{Mehdi Mirza}, \bibinfo{person}{Bing Xu},
  \bibinfo{person}{David Warde-Farley}, \bibinfo{person}{Sherjil Ozair},
  \bibinfo{person}{Aaron Courville}, {and} \bibinfo{person}{Yoshua Bengio}.}
  \bibinfo{year}{2014}\natexlab{}.
\newblock \showarticletitle{Generative adversarial nets}. In
  \bibinfo{booktitle}{{\em Advances in neural information processing systems}}.
  \bibinfo{pages}{2672--2680}.
\newblock


\bibitem[\protect\citeauthoryear{Goyal, Khot, Agrawal, Summers-Stay, Batra, and
  Parikh}{Goyal et~al\mbox{.}}{2018}]%
        {Goyal2018}
\bibfield{author}{\bibinfo{person}{Yash Goyal}, \bibinfo{person}{Tejas Khot},
  \bibinfo{person}{Aishwarya Agrawal}, \bibinfo{person}{Douglas Summers-Stay},
  \bibinfo{person}{Dhruv Batra}, {and} \bibinfo{person}{Devi Parikh}.}
  \bibinfo{year}{2018}\natexlab{}.
\newblock \showarticletitle{Making the V in VQA Matter: Elevating the Role of
  Image Understanding in Visual Question Answering}.
\newblock \bibinfo{journal}{{\em International Journal of Computer Vision\/}}
  (\bibinfo{year}{2018}).
\newblock


\bibitem[\protect\citeauthoryear{He, Zhang, Ren, and Sun}{He
  et~al\mbox{.}}{2016}]%
        {he2016deep}
\bibfield{author}{\bibinfo{person}{Kaiming He}, \bibinfo{person}{Xiangyu
  Zhang}, \bibinfo{person}{Shaoqing Ren}, {and} \bibinfo{person}{Jian Sun}.}
  \bibinfo{year}{2016}\natexlab{}.
\newblock \showarticletitle{Deep residual learning for image recognition}. In
  \bibinfo{booktitle}{{\em Proceedings of the IEEE conference on computer
  vision and pattern recognition}}. \bibinfo{pages}{770--778}.
\newblock


\bibitem[\protect\citeauthoryear{Jaderberg, Simonyan, Vedaldi, and
  Zisserman}{Jaderberg et~al\mbox{.}}{2016}]%
        {Jaderberg2016}
\bibfield{author}{\bibinfo{person}{Max Jaderberg}, \bibinfo{person}{Karen
  Simonyan}, \bibinfo{person}{Andrea Vedaldi}, {and} \bibinfo{person}{Andrew
  Zisserman}.} \bibinfo{year}{2016}\natexlab{}.
\newblock \showarticletitle{Reading Text in the Wild with Convolutional Neural
  Networks}.
\newblock \bibinfo{journal}{{\em Int. J. Comput. Vision\/}}
  \bibinfo{volume}{116}, \bibinfo{number}{1} (\bibinfo{year}{2016}),
  \bibinfo{pages}{1--20}.
\newblock


\bibitem[\protect\citeauthoryear{Karpathy and Fei-Fei}{Karpathy and
  Fei-Fei}{2017}]%
        {Karpathy17}
\bibfield{author}{\bibinfo{person}{A. Karpathy} {and} \bibinfo{person}{L.
  Fei-Fei}.} \bibinfo{year}{2017}\natexlab{}.
\newblock \showarticletitle{Deep Visual-Semantic Alignments for Generating
  Image Descriptions}.
\newblock \bibinfo{journal}{{\em IEEE Transactions on Pattern Analysis and
  Machine Intelligence\/}} \bibinfo{volume}{39}, \bibinfo{number}{4}
  (\bibinfo{year}{2017}), \bibinfo{pages}{664--676}.
\newblock


\bibitem[\protect\citeauthoryear{Kasturi, O'Gorman, and Govindaraju}{Kasturi
  et~al\mbox{.}}{2002}]%
        {Kasturi2002}
\bibfield{author}{\bibinfo{person}{Rangachar Kasturi},
  \bibinfo{person}{Lawrence O'Gorman}, {and} \bibinfo{person}{Venu
  Govindaraju}.} \bibinfo{year}{2002}\natexlab{}.
\newblock \showarticletitle{Document image analysis: A primer}.
\newblock \bibinfo{journal}{{\em Sadhana\/}} \bibinfo{volume}{27},
  \bibinfo{number}{1} (\bibinfo{year}{2002}), \bibinfo{pages}{3--22}.
\newblock


\bibitem[\protect\citeauthoryear{Kembhavi, Salvato, Kolve, Seo, Hajishirzi, and
  Farhadi}{Kembhavi et~al\mbox{.}}{2016}]%
        {Kembhavi2016}
\bibfield{author}{\bibinfo{person}{Aniruddha Kembhavi}, \bibinfo{person}{Mike
  Salvato}, \bibinfo{person}{Eric Kolve}, \bibinfo{person}{Minjoon Seo},
  \bibinfo{person}{Hannaneh Hajishirzi}, {and} \bibinfo{person}{Ali Farhadi}.}
  \bibinfo{year}{2016}\natexlab{}.
\newblock \showarticletitle{A Diagram is Worth a Dozen Images}. In
  \bibinfo{booktitle}{{\em European Conference on Computer Vision (ECCV)}},
  \bibfield{editor}{\bibinfo{person}{Bastian Leibe}, \bibinfo{person}{Jiri
  Matas}, \bibinfo{person}{Nicu Sebe}, {and} \bibinfo{person}{Max Welling}}
  (Eds.). \bibinfo{pages}{235--251}.
\newblock


\bibitem[\protect\citeauthoryear{Kingma and Ba}{Kingma and Ba}{2015}]%
        {Kingma2015AdamAM}
\bibfield{author}{\bibinfo{person}{Diederik~P. Kingma} {and}
  \bibinfo{person}{Jimmy Ba}.} \bibinfo{year}{2015}\natexlab{}.
\newblock \showarticletitle{Adam: A Method for Stochastic Optimization}.
\newblock \bibinfo{journal}{{\em CoRR\/}}  \bibinfo{volume}{abs/1412.6980}
  (\bibinfo{year}{2015}).
\newblock


\bibitem[\protect\citeauthoryear{Li, Adelson, and Agarwala}{Li
  et~al\mbox{.}}{2008}]%
        {li2008scribbleboost}
\bibfield{author}{\bibinfo{person}{Yuanzhen Li}, \bibinfo{person}{Edward
  Adelson}, {and} \bibinfo{person}{Aseem Agarwala}.}
  \bibinfo{year}{2008}\natexlab{}.
\newblock \showarticletitle{ScribbleBoost: Adding Classification to Edge-Aware
  Interpolation of Local Image and Video Adjustments}. In
  \bibinfo{booktitle}{{\em Computer Graphics Forum}},
  Vol.~\bibinfo{volume}{27}. Wiley Online Library, \bibinfo{pages}{1255--1264}.
\newblock


\bibitem[\protect\citeauthoryear{Liu, Reda, Shih, Wang, Tao, and Catanzaro}{Liu
  et~al\mbox{.}}{2018}]%
        {liu2018image}
\bibfield{author}{\bibinfo{person}{Guilin Liu}, \bibinfo{person}{Fitsum~A
  Reda}, \bibinfo{person}{Kevin~J Shih}, \bibinfo{person}{Ting-Chun Wang},
  \bibinfo{person}{Andrew Tao}, {and} \bibinfo{person}{Bryan Catanzaro}.}
  \bibinfo{year}{2018}\natexlab{}.
\newblock \showarticletitle{Image inpainting for irregular holes using partial
  convolutions}.
\newblock \bibinfo{journal}{{\em arXiv preprint arXiv:1804.07723\/}}
  (\bibinfo{year}{2018}).
\newblock


\bibitem[\protect\citeauthoryear{Lu, Wang, Lanir, Zhao, Pfister, Cohen-Or, and
  Huang}{Lu et~al\mbox{.}}{2020}]%
        {Lu2020}
\bibfield{author}{\bibinfo{person}{Min Lu}, \bibinfo{person}{Chufeng Wang},
  \bibinfo{person}{Joel Lanir}, \bibinfo{person}{Nanxuan Zhao},
  \bibinfo{person}{Hanspeter Pfister}, \bibinfo{person}{Daniel Cohen-Or}, {and}
  \bibinfo{person}{Hui Huang}.} \bibinfo{year}{2020}\natexlab{}.
\newblock \showarticletitle{Exploring Visual Information Flows in
  Infographics}. In \bibinfo{booktitle}{{\em Proceedings of the 2020 CHI
  Conference on Human Factors in Computing Systems}} {\em
  (\bibinfo{series}{CHI})}. \bibinfo{pages}{1–12}.
\newblock


\bibitem[\protect\citeauthoryear{Madan, Bylinskii, Tancik, Recasens, Zhong,
  Alsheikh, Pfister, Oliva, and Durand}{Madan et~al\mbox{.}}{2018}]%
        {visually2}
\bibfield{author}{\bibinfo{person}{Spandan Madan}, \bibinfo{person}{Zoya
  Bylinskii}, \bibinfo{person}{Matthew Tancik}, \bibinfo{person}{Adrià
  Recasens}, \bibinfo{person}{Kimberli Zhong}, \bibinfo{person}{Sami Alsheikh},
  \bibinfo{person}{Hanspeter Pfister}, \bibinfo{person}{Aude Oliva}, {and}
  \bibinfo{person}{Fredo Durand}.} \bibinfo{year}{2018}\natexlab{}.
\newblock \showarticletitle{Synthetically Trained Icon Proposals for Parsing
  and Summarizing Infographics}. In \bibinfo{booktitle}{{\em arXiv preprint
  arXiv:1807.10441}}.
\newblock
\showURL{%
\url{https://arxiv.org/pdf/1807.10441}}


\bibitem[\protect\citeauthoryear{Meyer, Cornill{\`{e}}re, Djelouah, Schroers,
  and Gross}{Meyer et~al\mbox{.}}{2018}]%
        {DeepVideoPropagation}
\bibfield{author}{\bibinfo{person}{Simone Meyer}, \bibinfo{person}{Victor
  Cornill{\`{e}}re}, \bibinfo{person}{Abdelaziz Djelouah},
  \bibinfo{person}{Christopher Schroers}, {and} \bibinfo{person}{Markus~H.
  Gross}.} \bibinfo{year}{2018}\natexlab{}.
\newblock \showarticletitle{Deep Video Color Propagation}. In
  \bibinfo{booktitle}{{\em British Machine Vision Conference, {BMVC}}}.
  \bibinfo{pages}{128}.
\newblock


\bibitem[\protect\citeauthoryear{O'Gorman}{O'Gorman}{1993}]%
        {Ogorman1993}
\bibfield{author}{\bibinfo{person}{Lawrence O'Gorman}.}
  \bibinfo{year}{1993}\natexlab{}.
\newblock \showarticletitle{The document spectrum for page layout analysis}.
\newblock \bibinfo{journal}{{\em IEEE Transactions on Pattern Analysis and
  Machine Intelligence\/}} \bibinfo{volume}{15}, \bibinfo{number}{11}
  (\bibinfo{year}{1993}), \bibinfo{pages}{1162--1173}.
\newblock


\bibitem[\protect\citeauthoryear{P{\'e}rez, Gangnet, and Blake}{P{\'e}rez
  et~al\mbox{.}}{2003}]%
        {perez2003poisson}
\bibfield{author}{\bibinfo{person}{Patrick P{\'e}rez}, \bibinfo{person}{Michel
  Gangnet}, {and} \bibinfo{person}{Andrew Blake}.}
  \bibinfo{year}{2003}\natexlab{}.
\newblock \showarticletitle{Poisson image editing}.
\newblock \bibinfo{journal}{{\em ACM Transactions on graphics (TOG)\/}}
  \bibinfo{volume}{22}, \bibinfo{number}{3} (\bibinfo{year}{2003}),
  \bibinfo{pages}{313--318}.
\newblock


\bibitem[\protect\citeauthoryear{Poco and Heer}{Poco and Heer}{2017}]%
        {Poco2017}
\bibfield{author}{\bibinfo{person}{Jorge Poco} {and} \bibinfo{person}{Jeffrey
  Heer}.} \bibinfo{year}{2017}\natexlab{}.
\newblock \showarticletitle{Reverse-Engineering Visualizations: Recovering
  Visual Encodings from Chart Images}.
\newblock \bibinfo{journal}{{\em Computer Graphics Forum\/}}
  \bibinfo{volume}{36}, \bibinfo{number}{3} (\bibinfo{year}{2017}),
  \bibinfo{pages}{353--363}.
\newblock


\bibitem[\protect\citeauthoryear{Saund, Fleet, Larner, and Mahoney}{Saund
  et~al\mbox{.}}{2003}]%
        {Saund2003}
\bibfield{author}{\bibinfo{person}{Eric Saund}, \bibinfo{person}{David Fleet},
  \bibinfo{person}{Daniel Larner}, {and} \bibinfo{person}{James Mahoney}.}
  \bibinfo{year}{2003}\natexlab{}.
\newblock \showarticletitle{Perceptually-supported Image Editing of Text and
  Graphics}. In \bibinfo{booktitle}{{\em Proceedings of the 16th Annual ACM
  Symposium on User Interface Software and Technology}}.
  \bibinfo{pages}{183--192}.
\newblock


\bibitem[\protect\citeauthoryear{Shah and Koltun}{Shah and Koltun}{2018}]%
        {Shah2018}
\bibfield{author}{\bibinfo{person}{Sohil~Atul Shah} {and}
  \bibinfo{person}{Vladlen Koltun}.} \bibinfo{year}{2018}\natexlab{}.
\newblock \showarticletitle{Deep Continuous Clustering}.
\newblock \bibinfo{journal}{{\em CoRR\/}} (\bibinfo{year}{2018}).
\newblock


\bibitem[\protect\citeauthoryear{Shen, Hertzmann, Jia, Paris, Price, Shechtman,
  and Sachs}{Shen et~al\mbox{.}}{2016}]%
        {shen2016automatic}
\bibfield{author}{\bibinfo{person}{Xiaoyong Shen}, \bibinfo{person}{Aaron
  Hertzmann}, \bibinfo{person}{Jiaya Jia}, \bibinfo{person}{Sylvain Paris},
  \bibinfo{person}{Brian Price}, \bibinfo{person}{Eli Shechtman}, {and}
  \bibinfo{person}{Ian Sachs}.} \bibinfo{year}{2016}\natexlab{}.
\newblock \showarticletitle{Automatic portrait segmentation for image
  stylization}. In \bibinfo{booktitle}{{\em Computer Graphics Forum}},
  Vol.~\bibinfo{volume}{35}. Wiley Online Library, \bibinfo{pages}{93--102}.
\newblock


\bibitem[\protect\citeauthoryear{Smith}{Smith}{2007}]%
        {smith2007overview}
\bibfield{author}{\bibinfo{person}{Ray Smith}.}
  \bibinfo{year}{2007}\natexlab{}.
\newblock \showarticletitle{An overview of the Tesseract OCR engine}. In
  \bibinfo{booktitle}{{\em Ninth International Conference on Document Analysis
  and Recognition (ICDAR 2007)}}, Vol.~\bibinfo{volume}{2}.
  \bibinfo{pages}{629--633}.
\newblock


\bibitem[\protect\citeauthoryear{Xie, Girshick, and Farhadi}{Xie
  et~al\mbox{.}}{2016}]%
        {xie2016unsupervised}
\bibfield{author}{\bibinfo{person}{Junyuan Xie}, \bibinfo{person}{Ross
  Girshick}, {and} \bibinfo{person}{Ali Farhadi}.}
  \bibinfo{year}{2016}\natexlab{}.
\newblock \showarticletitle{Unsupervised deep embedding for clustering
  analysis}. In \bibinfo{booktitle}{{\em International conference on machine
  learning}}. \bibinfo{pages}{478--487}.
\newblock


\bibitem[\protect\citeauthoryear{Xing, Chen, and Wei}{Xing
  et~al\mbox{.}}{2014}]%
        {AutocompletePainting}
\bibfield{author}{\bibinfo{person}{Jun Xing}, \bibinfo{person}{Hsiang-Ting
  Chen}, {and} \bibinfo{person}{Li-Yi Wei}.} \bibinfo{year}{2014}\natexlab{}.
\newblock \showarticletitle{Autocomplete Painting Repetitions}.
\newblock \bibinfo{journal}{{\em ACM Trans. Graph.\/}} \bibinfo{volume}{33},
  \bibinfo{number}{6} (\bibinfo{year}{2014}), \bibinfo{pages}{172:1--172:11}.
\newblock


\bibitem[\protect\citeauthoryear{Xu, Li, Ju, Hu, and Liu}{Xu
  et~al\mbox{.}}{2009}]%
        {xu2009efficient}
\bibfield{author}{\bibinfo{person}{Kun Xu}, \bibinfo{person}{Yong Li},
  \bibinfo{person}{Tao Ju}, \bibinfo{person}{Shi-Min Hu}, {and}
  \bibinfo{person}{Tian-Qiang Liu}.} \bibinfo{year}{2009}\natexlab{}.
\newblock \showarticletitle{Efficient affinity-based edit propagation using kd
  tree}. In \bibinfo{booktitle}{{\em ACM Transactions on Graphics (TOG)}},
  Vol.~\bibinfo{volume}{28}. ACM, \bibinfo{pages}{118}.
\newblock


\bibitem[\protect\citeauthoryear{Yang, Parikh, and Batra}{Yang
  et~al\mbox{.}}{2016}]%
        {yang2016joint}
\bibfield{author}{\bibinfo{person}{Jianwei Yang}, \bibinfo{person}{Devi
  Parikh}, {and} \bibinfo{person}{Dhruv Batra}.}
  \bibinfo{year}{2016}\natexlab{}.
\newblock \showarticletitle{Joint unsupervised learning of deep representations
  and image clusters}. In \bibinfo{booktitle}{{\em Proceedings of the IEEE
  Conference on Computer Vision and Pattern Recognition}}.
  \bibinfo{pages}{5147--5156}.
\newblock


\bibitem[\protect\citeauthoryear{Yang, Yumer, Asente, Kraley, Kifer, and
  Lee~Giles}{Yang et~al\mbox{.}}{2017}]%
        {yang2017learning}
\bibfield{author}{\bibinfo{person}{Xiao Yang}, \bibinfo{person}{Ersin Yumer},
  \bibinfo{person}{Paul Asente}, \bibinfo{person}{Mike Kraley},
  \bibinfo{person}{Daniel Kifer}, {and} \bibinfo{person}{C Lee~Giles}.}
  \bibinfo{year}{2017}\natexlab{}.
\newblock \showarticletitle{Learning to Extract Semantic Structure From
  Documents Using Multimodal Fully Convolutional Neural Networks}. In
  \bibinfo{booktitle}{{\em Proceedings of the IEEE Conference on Computer
  Vision and Pattern Recognition}}. \bibinfo{pages}{5315--5324}.
\newblock


\bibitem[\protect\citeauthoryear{Zhu, Park, Isola, and Efros}{Zhu
  et~al\mbox{.}}{2017}]%
        {zhu2017unpaired}
\bibfield{author}{\bibinfo{person}{Jun-Yan Zhu}, \bibinfo{person}{Taesung
  Park}, \bibinfo{person}{Phillip Isola}, {and} \bibinfo{person}{Alexei~A
  Efros}.} \bibinfo{year}{2017}\natexlab{}.
\newblock \showarticletitle{Unpaired Image-to-Image Translation using
  Cycle-Consistent Adversarial Networks}. In \bibinfo{booktitle}{{\em IEEE
  International Conference on Computer Vision}}.
\newblock


\end{thebibliography}
